\documentclass[]{template}

\usepackage[utf8]{inputenc}             
\usepackage[T1]{fontenc}                
\usepackage{url}                        
\usepackage{booktabs}                   
\usepackage{multirow}
\usepackage{colortbl}
\usepackage{multicol}
\usepackage{amsfonts}                   
\usepackage{nicefrac}                   
\usepackage{microtype}                  
\usepackage[dvipsnames]{xcolor}         

\usepackage{latexsym}

\usepackage{graphicx}
\usepackage{float}

\usepackage{bm}

\usepackage{tabularx} 
\usepackage{ragged2e} 
\newcolumntype{L}{>{\RaggedRight\hangafter=1\hangindent=0em}X}

\usepackage{enumitem}

\usepackage{amsmath}
\usepackage{amssymb}
\usepackage{mathtools}
\usepackage{amsthm}

\usepackage{subfigure}

\usepackage{booktabs} 
\usepackage{array}    
\usepackage{caption}  
\usepackage{multirow}

\usepackage{inconsolata}
\usepackage{algorithmic}
\usepackage{hyperref}

\setboolean{logo}{true}    

\usepackage[linesnumbered,ruled,vlined]{algorithm2e}

\hypersetup{
    colorlinks=true,
    linkcolor=red,
    citecolor=Cerulean,
    filecolor=magenta,      
    urlcolor=magenta,
}

\usepackage[capitalize,noabbrev]{cleveref}
\crefname{section}{§}{§§}
\Crefname{section}{§}{§§}

\usepackage{calligra}
\DeclareMathAlphabet{\mathcalligra}{T1}{calligra}{m}{n}

\usepackage{pifont}

\theoremstyle{plain}

\theoremstyle{definition}

\theoremstyle{remark}

\renewcommand{\paragraph}[1]{\vspace{1mm}\noindent\textbf{#1}}

\DeclareCaptionLabelFormat{cont}{#1~#2\alph{ContinuedFloat}}
\usepackage[most]{tcolorbox}
\tcbset{
  promptbox/.style={
    top=10pt,
    colback=lightgray!20,
    colframe=Black,
    colbacktitle=NavyBlue,
    enhanced,
    center,
    attach boxed title to top center={yshift=-0.1in,xshift=0.0in},
    boxed title style={boxrule=0pt,colframe=white,},
  }
}
\newtcolorbox{promptbox}[2][]{promptbox, title=#2,#1}
\tcbset{
  takeawaybox/.style={
    top=10pt,
    colback=lightgray!20,
    colframe=Black,
    colbacktitle=BurntOrange,
    enhanced,
    center,
    attach boxed title to top center={yshift=-0.1in,xshift=0.0in},
    boxed title style={boxrule=0pt,colframe=white,},
  }
}
\newtcolorbox{takeawaybox}[2][]{takeawaybox, title=#2,#1}
\tcbset{
  observationbox/.style={
    top=10pt,
    colback=lightgray!20,
    colframe=Black,
    colbacktitle=YellowGreen,
    enhanced,
    center,
    attach boxed title to top center={yshift=-0.1in,xshift=0.0in},
    boxed title style={boxrule=0pt,colframe=white,},
  }
}
\newtcolorbox{observationbox}[2][]{observationbox, title=#2,#1}

\usepackage{xspace}

\newcommand\blfootnote[1]{%
  \begingroup
  \renewcommand\thefootnote{}\footnote{#1}%
  \addtocounter{footnote}{-1}%
  \endgroup
}

\usepackage{CJK}

\title{How to Set the Learning Rate for Large-Scale Pre-training?}

\author{%
    \textbf{Yunhua Zhou}${}^{1*}$ \quad
    \textbf{Shuhao Xing}${}^{3,1*}$ \quad
    \textbf{Junhao Huang}${}^{2,1}$ \quad
    \textbf{Xipeng Qiu}${}^{3\dag}$ \quad
    \textbf{Qipeng Guo}${}^{1\dag}$ \\
    \textsuperscript{1}Shanghai AI Laboratory, China \\
    \textsuperscript{2}Shanghai JiaoTong University, China \\
    \textsuperscript{3}Fudan University, China \\
    \texttt{\{zhouyunhua, xingshuhao.dispatch\}@pjlab.org.cn}
}%

\affil[1]{Shanghai AI Laboratory}

\begin{abstract}
Optimal configuration of the learning rate (LR) is a fundamental yet formidable challenge in large-scale pre-training. Given the stringent trade-off between training costs and model performance, the pivotal question is whether the optimal LR can be accurately extrapolated from low-cost experiments. In this paper, we formalize this investigation into two distinct research paradigms: Fitting and Transfer. Within the Fitting Paradigm, we innovatively introduce a Scaling Law for search factor, effectively reducing the search complexity from $\mathcal{O}(n^3)$ to $\mathcal{O}(n \cdot C_{D} \cdot C_{\eta})$ via predictive modeling. Within the Transfer Paradigm, we extend the principles of $\mu$Transfer to the Mixture of Experts (MoE) architecture, broadening its applicability to encompass model depth, weight decay, and token horizons.

By pushing the boundaries of existing hyperparameter research in terms of scale, we conduct a comprehensive comparison between these two paradigms. Our empirical results challenge the scalability of the widely adopted $\mu$Transfer in large-scale pre-training scenarios. Furthermore, we provide a rigorous analysis through the dual lenses of training stability and feature learning to elucidate the underlying reasons why module-wise parameter tuning underperforms in large-scale settings. This work offers systematic practical guidelines and a fresh theoretical perspective for optimizing industrial-level pre-training.

\end{abstract}

\begin{document}

\blfootnote{$*$ Equal contribution. Orders are determined randomly.}
\blfootnote{$\dagger$ Corresponding author.}

\maketitle

\section{Introduction}
The rapid evolution of Large Language Models (LLMs)~\cite{openai2024gpt4technicalreport,openai2024gpt4ocard,openai2024openaio1card,openai2025gptoss120bgptoss20bmodel,deepseekai2024deepseekllmscalingopensource,deepseekai2024deepseekv2strongeconomicalefficient,deepseekv3,deepseekai2025deepseekr1incentivizingreasoningcapability,deepseekai2025deepseekv32pushingfrontieropen} is continuously pushing the cognitive boundaries of artificial intelligence, driven fundamentally by the Scaling Laws~\cite{kaplan-scalinglaw} arising from large-scale pre-training. However, executing such large-scale pre-training remains formidable, A fundamental challenge is selecting an appropriate/optimal learning rate (LR). On one hand, large-scale pre-training involves massive computational loads and prolonged training cycles, requiring a precise LR to ensure both stability and convergence efficiency. On any other hand, the vast consumption of computational resources makes the cost of trial-and-error unacceptable. Consequently, \textbf{the crux of learning rate for large-scale pre-training lies in accurately characterizing the relationship between the optimal LR in ``cheaper-to-train'' small-scale experiments and that of the target scale.}

This paper establishes two fundamental research paradigms for setting the learning rate in large-scale pre-training: \textbf{Fitting} and \textbf{Transfer}.
The Fitting Paradigm involves directly modeling the relationship between the optimal learning rate, model size, and training data under standard initialization conditions, thereby extrapolating the learning rate for the target training scale~\cite{deepseekai2024deepseekllmscalingopensource,steplaw}. To overcome the bottlenecks of combinatorial explosion and prohibitive training costs inherent in prior research within the fitting paradigm, this work innovatively introduces a scaling Law for search factor. By leveraging performance prediction, we effectively reduce the search complexity from $\mathcal{O}(n^3)$ to $\mathcal{O}(n \cdot C_{D} \cdot C_{\eta})$.

The Transfer Paradigm, on the other hand, conducts hyperparameter optimization (including learning rate) on selected proxy models and subsequently transfers these hyperparameters to the target model according to established rules. In this study, we adopt the fundamental principles of $\mu$Transfer~\cite{mup}. However, to better align with contemporary large-scale pre-training scenarios, we implement a critical extension by selecting the Mixture of Experts (MoE) as our research architecture. Building upon existing literature, we expand the transfer dimensions to encompass model widths and depths, while simultaneously incorporating the influences of weight decay and token horizon on $\mu$Transfer. These enhancements substantially push the boundary of applicability for $\mu$Transfer.

To facilitate a comprehensive comparison between the two paradigms, this study extends the target prediction scale for learning rates by more than tenfold (x10). Our target configuration is set as a MoE model with 12B total parameters with 1.3B activated for each token, trained on 500B tokens—a scale that significantly surpasses existing hyperparameter research. The primary contributions of this paper can be summarized in the following three aspects:\\
\textbf{Paradigm and Theoretical Innovation:} We systematically formalize the two research paradigms and innovatively integrate Scaling Laws for performance prediction. This approach effectively reduces modeling costs while substantially enhancing both prediction efficiency and the range of parameter coverage.\\
\textbf{Ultra-Large-Scale Empirical Comparison:} Breaking through the scale limitations of prior hyperparameter studies, this work provides the first comprehensive comparison of the two paradigms within a real-world, large-scale pre-training environment, offering systematic practical guidelines for large-model engineering.\\
\textbf{Multidimensional Mechanistic Insights:} We provide an in-depth analysis of the dynamical characteristics of both paradigms during pre-training, focusing on two core dimensions: Training Stability and Feature Learning. This offers a novel perspective for research into large-scale pre-training.\\

\section{Related Works}

\subsection{Learning Rate Schedule}

Prior to the advent of large-scale language model (LLM) pre-training, the cosine annealing schedule \cite{cosineschedule} served as the predominant standard. However, the cosine schedule mandates a predetermined number of total training steps, rendering it insufficiently flexible amidst the backdrop of continuously expanding pre-training scales. Consequently, the Warmup-Stable-Decay (WSD) schedule \cite{minicpm} has emerged. This schedule is characterized by a stable phase where the learning rate remains constant following the warmup period, eventually decaying to a specific terminal value. Since the decay phase can be initiated at any point during the stable phase to conclude training, WSD is regarded as highly adaptable to the dynamic requirements of large-scale pre-training. Reflecting this advantage, the WSD scheduler has recently been adopted by mainstream large-scale pre-training projects\cite{deepseekv3,kimik2,interns1}.

Propelled by scaling laws, the magnitude of pre-training continues to escalate. The stable phase frequently spans weeks or even months \cite{deepseekv3,interns1,yang2025qwen3technicalreport}, making the precise configuration of the learning rate critically important. However, existing research on learning rate configuration has predominantly focused on the cosine annealing schedule. Under the cosine regime, \citet{kaplan-scalinglaw} elucidated the relationship between the learning rate and model parameters, while \citet{bjorck-scalinglaw} and \citet{steplaw} empirically derived power-law formulations correlating the learning rate with model size $N$ and training data size $D$. Diverging from existing literature, our work investigates the relationship between the optimal learning rate, model size, and training data size specifically within the stable phase of a constant learning rate schedule.

\subsection{Maximal Update  Parametrization}

Maximal Update  Parametrization ($\mu$Parametrization or $\mu$P, \citet{mup}) is a widely investigated framework for hyperparameter configuration. The fundamental premise of $\mu$P is to guarantee training stability and ensure that weights across different modules are adequately trained(i.e. maximal feature learning) even as model width approaches infinity.

By virtue of maintaining these properties in the infinite-width limit, $\mu$P possesses inherent capabilities for hyperparameter transfer. This gives rise to a derivative method known as $\mu$Transfer, wherein the optimal learning rate for a target model can be directly calculated based on the optimum identified via search on a smaller proxy model. While the initial formulation of $\mu$Transfer was limited to extrapolating model width, subsequent studies by \citet{depthmup} and \citet{dey2025dontlazycompletepenables} have investigated extensions for scaling model depth. Beyond its extensive application in dense architectures\cite{empiricalstudymuplearning}, $\mu$Transfer has also been experimentally applied to Mixture-of-Experts (MoE) structures\cite{moemup}. Furthermore, recent research indicates that the efficacy of $\mu$Transfer is primarily manifested during the early stages of training; to extend the effective transfer horizon, adjustments to weight decay are required\cite{wang2025setadamwsweightdecay,moemup,fan2025robustlayerwisescalingrules}.

Building upon existing research of $\mu$Transfer and integrating current methodologies for pre-training hyperparameter configuration, our work conducts a granular investigation into the impact of $\mu$Transfer on the performance of large-scale pre-training.

\section{Approach}

This section delineates the specific methodologies for the configuration of the learning rate under two distinct paradigms. Section \ref{sec:lr-scaling-law} introduces the Fitting Paradigm, which leverages scaling laws to enhance the efficiency and scope of the fitting process. Section \ref{sec:mup} focuses on the representative transfer paradigm $\mu$Transfer method and elucidating its practical implementation within large-scale pre-training contexts. Crucially, our study focuses on the stable training phase governed by the Warmup-Stable-Decay (WSD) learning rate schedule.

\subsection{Scaling Laws for Learning Rate}
\label{sec:lr-scaling-law}

For a given model size $N$ and training data size $D$ , the optimal learning rate $\eta$ is formulated as:

\begin{equation}
    \eta^{*}_{ND} = \operatorname*{argmin}_{\eta} L(\eta \mid N, D, \Theta),
    \label{eq:opt-lr-def}
\end{equation}

where L is validation loss and $\Theta$ contains other hyperparameters involved in the pre-train process.

Characterizing the relationship between the optimal $\eta$, model size, and data size necessitates a grid search across the $N$, $D$, and $\eta$ dimensions, resulting in a computational complexity of approximately $\mathcal{O}(n^3)$. Fortunately, inspired by prior work \cite{bjorck-scalinglaw}, we observe that for fixed $N$ and $D$, the relationship between the validation loss $L$ and the learning rate $\eta$ approximates an invex profile. Consequently, we employ a quadratic polynomial to fit this relationship:

\begin{equation}
    L(\eta \mid N, D, \Theta) = L_{min} + C \cdot (\log(\eta) - \eta_{min})^2,
    \label{eq:loss-n-d}
\end{equation}

where $L_{min}$, $C$, and $\eta_{min}$ are the fitting coefficients.

Consequently, for a given $N$ and $D$, the optimal learning rate $\eta^*$ can be directly derived via fitting on a limited set of learning rates:

\begin{equation}
    \log(\eta^{*}) = \eta_{min} = \operatorname*{argmin}_{\eta} \{ L(\eta \mid N, D, \Theta) \},
    \label{eq:log-lr-n-d}
\end{equation}

\begin{figure}
    \centering
    \vspace{-0.35cm}
    \subfigtopskip=2pt
    \subfigbottomskip=2pt
    \subfigcapskip=-5pt
    \subfigure[Fitting results for Equation \ref{eq:ld}. Data points to the left of the dashed line represent the empirical values used for fitting; The curves to the right side depicts predictions. See Appendix\ref{sec:appendix-ld} for the discussion on the accuracy of Equation \ref{eq:ld}.]
    {
        \label{fig:ld-exm}
        \includegraphics[width=0.43\linewidth]{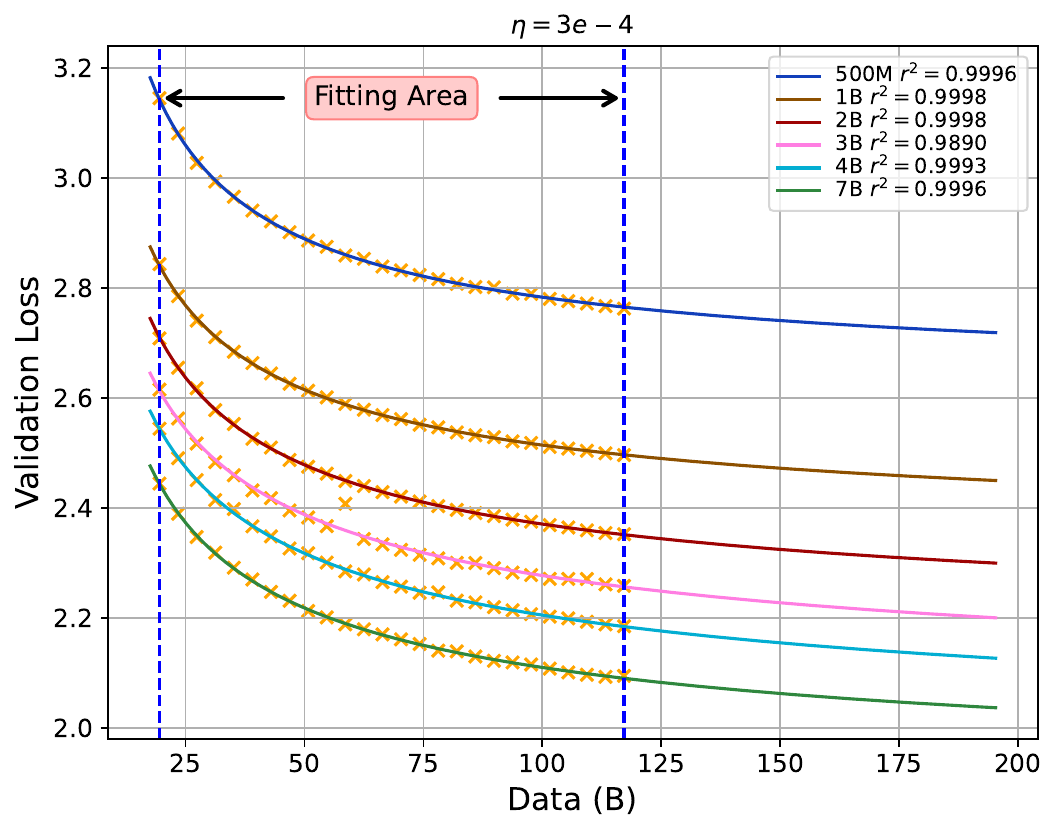}
    }
    \quad
    \subfigure[Results of fitting the validation loss against the learning rate (LR) using a \textbf{quadratic polynomial}. Different colored curves correspond to models of varying sizes, while the triangle indicate the optimal LR.]
    {
        \label{fig:l-lr-exm}
        \includegraphics[width=0.43\linewidth]{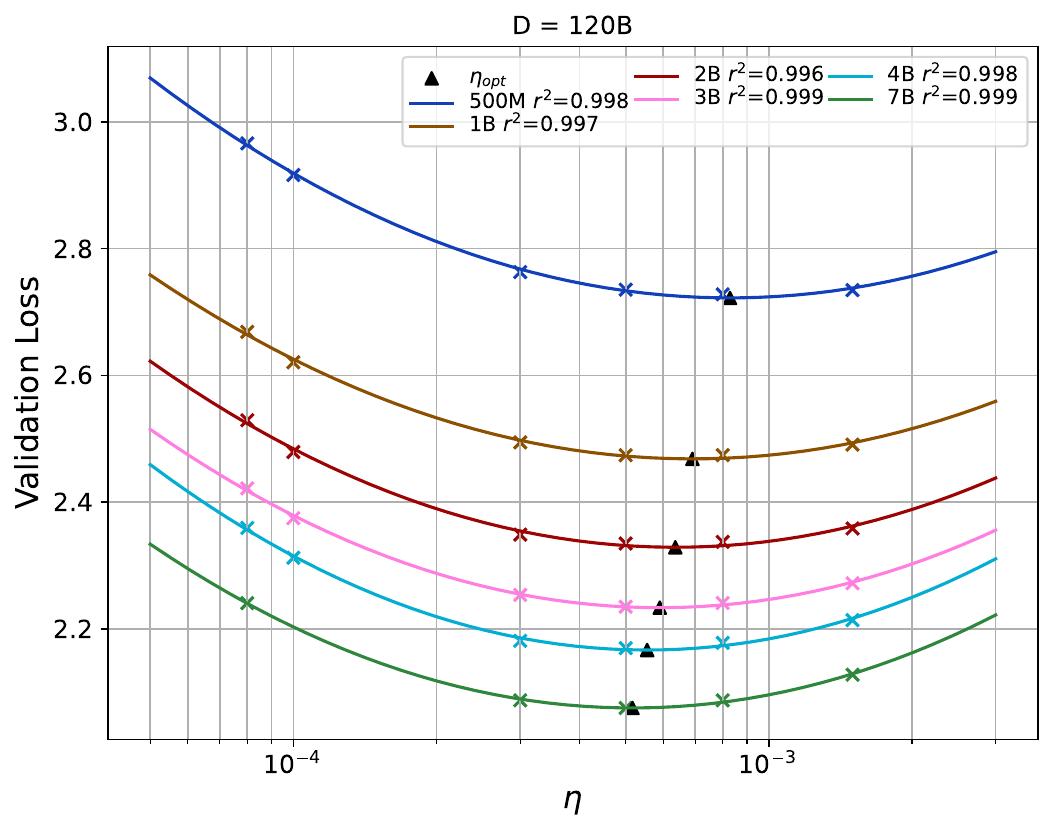}
    }
    \caption{Results of Equation \ref{eq:ld} and \ref{eq:loss-n-d}. These approaches allow for a substantial reduction in the time and storage cost of the search process. }
    \label{fig:main-exm}
\end{figure}

Figure \ref{fig:l-lr-exm} shows the fitted curves of Equation \ref{eq:loss-n-d}. The search complexity is reduced from $\mathcal{O}(n^3)$ to $\mathcal{O}(n^2*C_{\eta})$

Furthermore, inspired by contemporary research on scaling laws\cite{hoffmann2022trainingcomputeoptimallargelanguage,tissue2024scalinglawlearningrate}, we observe that under the WSD schedule, the validation loss exhibits a power-law relationship with the training data size $D$ for a fixed model configuration:

\begin{equation}
    L(D) = L_0 + A \cdot D^{-\gamma},
    \label{eq:ld}
\end{equation}

Where $L_0, A, \gamma$ are parameters to fit. This implies that the search space along the dimension of data size $D$ can be significantly compressed, enabling the extrapolation of results to larger data regimes via a limited number of search points. The specific fitting procedure is illustrated in Figure \ref{fig:ld-exm}. This methodology effectively improves the trade-off between search cost and fitting precision, reducing the computational complexity of the search from $\mathcal{O}(n^3)$ to $\mathcal{O}(n \cdot C_{D} \cdot C_{\eta})$.

Based on Equation \ref{eq:loss-n-d} and \ref{eq:log-lr-n-d}, by conducting a search within the $N$ dimension, we can efficiently derive a comprehensive set of optimal LR $\{\eta^*_{ND}\}_{N,D}$, corresponding to varying model sizes $N$ and data sizes $D$. This facilitates the fitting of the functional relationship between the optimal LR and the variables $N$ and $D$:

\begin{equation}
    Lr(N,D) = \operatorname*{argmin}_{Lr \in \mathcal{F}} L(Lr(N,D), \eta^{*}_{ND} \mid \Theta),
    \label{eq:lr-n-d-obj}
\end{equation}

where $\mathcal{F}$ represents the candidate function space, and $L$ denotes the metric function, which is Root Mean Squared Error (RMSE) in our work.

The final fitted relationship governing the optimal learning rate with respect to model size $N$ and data size $D$ is given by:

\begin{equation}
    Lr(N,D) = 38.4588 \cdot N^{-0.2219} \cdot D^{-0.3509}.
    \label{eq:lr-n-d-res}
\end{equation}

\begin{figure*}[t]
    \centering
    \includegraphics[width=1\linewidth]{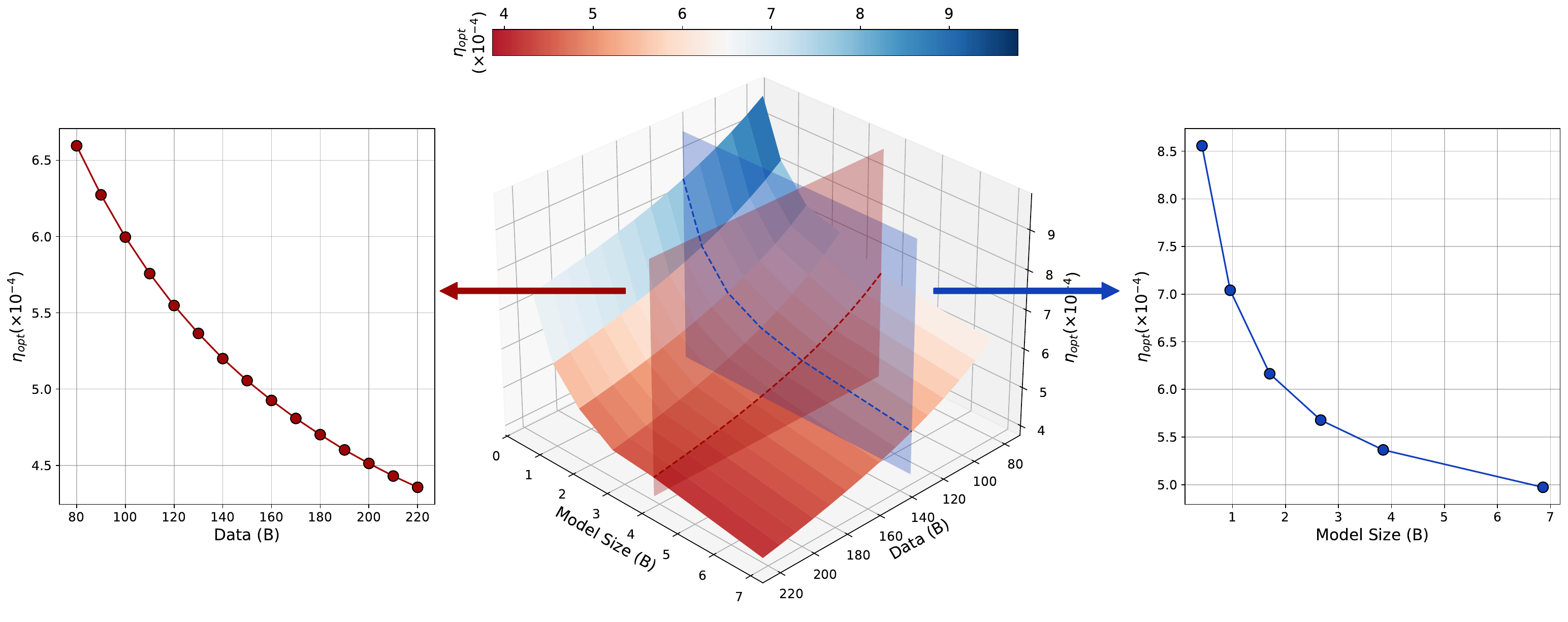}
    \caption{\textbf{Middle}: Visualization of the optimal learning rate relative to model size $N$ and data size $D$. \textbf{Left}: The relationship between the optimal learning rate and data size $D$ with model size fixed at $N=4\text{B}$. \textbf{Right}: The relationship between the optimal learning rate and model size $N$ with data size fixed at $D=140\text{B}$.}
    \label{fig:lr_n_d_final}
\end{figure*}

We observe a good fit with $R^2 \approx 0.9622$(See Appendix \ref{sec:appendix-global-lr} for details of fitting process). The overall fitting results are shown in Figure \ref{fig:lr_n_d_final}.

Extending this approach, we further conduct a fine-grained investigation into the learning rate configurations for distinct model modules in Section \ref{sec:module-lr}.

\subsection{Scaling $\mu$Transfer for Pre-training}
\label{sec:mup}

As the Mixture-of-Experts (MoE) architecture increasingly serves as the foundational backbone for large-scale pre-training \cite{deepseekai2024deepseekv2strongeconomicalefficient,deepseekv3,deepseekai2025deepseekr1incentivizingreasoningcapability,deepseekai2025deepseekv32pushingfrontieropen,yang2024qwen2technicalreport,qwen2025qwen25technicalreport,yang2025qwen3technicalreport,interns1}, we adopt the MoE architecture as our proxy model for $\mu$P. Regarding the target model, we adhere to the settings proposed by \citet{moemup} for initialization along the width dimension. For the depth dimension, we draw upon the methodologies of Depth-up \cite{depthmup} and Complete-$\mu$P \cite{dey2025dontlazycompletepenables,complete-d-p}. The central mechanism involves applying a depth-dependent scaling factor to the residual branch:

\begin{equation}
    H^{i+1} = H^{i} + m^{-\alpha}_{L}\mathcal{F}(H^{i}), \quad i \in \{1, \dots, L\},
\label{eq:depth-mup}
\end{equation}

where $H^i$ denotes the output of the $i$-th layer, and $\mathcal{F}$ represents either the Attention or Feed-Forward Network (FFN) layer. Following the recommendations of Complete-$\mu$P, we set $\alpha = 1$ to enhance the transferability of $\mu$Transfer.

\citet{wang2025setadamwsweightdecay} and \citet{fan2025robustlayerwisescalingrules} have identified weight decay $\lambda$ as a critical determinant of $\mu$Transfer efficacy. Consequently, we incorporate the influence of weight decay into the training process of the target model, maintaining the proportionality $\delta \lambda \propto \delta lr$. For given model size $N$ and data volume $D$, we observe that the approximate invex relationship between validation loss $L$ and learning rate persists within $\mu$P proxy models. This observation allows for a reduction in the search space along the learning rate dimension, thereby improving the efficiency of $\mu$Transfer. Regarding transfer along the token horizon dimension, we adopt the configuration from \citet{complete-d-p}. The detailed initialization and transfer rules for the target model parameters are summarized in Table \ref{tab:mup-models} and Table \ref{tab:mutransfer}.

\section{Experiments}

\subsection{Datasets}

The pre-training corpus utilized in our work is derived from InternLM2.5 \cite{cai2024internlm2technicalreport}, including general text, source code, and long-context sequences. Specifically, the textual component spans web pages, academic papers, patents, and books. The code component is primarily sourced from GitHub, programming communities, and other public repositories, covering a diverse array of programming languages including C/C++, Java, and Python. All data underwent rigorous deduplication and safety filtering protocols.

To ensure distributional consistency, the validation set employed in our experiments was constructed via random sampling from the above corpus, while strictly maintaining disjointness from the training samples to prevent data leakage.

\subsection{Experimental Settings}

We adopt the Qwen3-MoE \cite{yang2025qwen3technicalreport} architecture for our experimental models. For all model training, we utilize the AdamW optimizer \cite{adamw} with $\beta_1=0.9$, $\beta_2=0.95$, and $\epsilon=10^{-8}$. The learning rate schedule consists of a linear warmup for 1,000 steps, followed by a constant learning rate strategy. The sequence length is fixed at 4,096, and the global batch size is set to 4M tokens.

For experiments in \ref{sec:lr-scaling-law}, we employ models of four distinct sizes (550M, 1B, 2B, and 3B) all adhering to the structural configuration of the Qwen3-30B-A3B model. Notably, the aspect ratio between model width and depth remains constant across these scales. We subsequently validate our experimental findings on target models with 4B, 12B total parameters. With the exception of normalization parameters, all model weights are initialized from a normal distribution with a standard deviation of 0.02. The search space for the learning rate is defined as $\eta \in \{8e-5, 1e-4, 3e-4, 5e-4, 8e-4, 1.5e-3, 2e-3\}$. Each model is trained on approximately 120B tokens (30,000 steps), and we extrapolate the results to 500B tokens using Equation \ref{eq:ld}. Weight decay is set to 0.1.

For the $\mu$Transfer experiments, we conduct learning rate and initialization searches on a proxy model with 2B total parameters. The search space for the learning rate is defined as $\eta \in \{8e-5, 1e-4, 3e-4, 5e-4, 8e-4, 1.5e-3, 2e-3\}$; for initialization, we explore the range $\sigma \in \{0.0005, 0.001, 0.002, 0.005, 0.01, 0.015, 0.02\}$. The actual training data size for these experiments is approximately 200B tokens (50,000 steps), and we extrapolate the hyperparameters to a 500B token regime with Equation \ref{eq:ld} and settings from \citet{complete-d-p}.

\subsection{Evaluation}

To assess the downstream performance of the models developed during our validation experiments, we evaluate our models on MMLU \cite{mmlu}  and CMMLU \cite{li-etal-2024-cmmlu} benchmarks. MMLU serves as our primary English evaluation set, comprising four-choice multiple-choice questions across 57 distinct subjects, including anatomy, physics, genetics etc. Conversely, we employ CMMLU to evaluate Chinese language proficiency which covers 67 domains ranging from natural sciences and humanities to general knowledge.

For the implementation of these evaluations, we leverage the OpenCompass framework \cite{2023opencompass},
a comprehensive Python toolkit designed to facilitate the batch evaluation of diverse foundation models across heterogeneous datasets. Furthermore, to expedite the evaluation pipeline, we utilize the LMDeploy framework \cite{2023lmdeploy} with Turbomind \cite{zhang2025efficient} backend for efficient model loading and inference acceleration.

\section{Results}
\label{sec:results}

\begin{figure}
    \centering
    \includegraphics[width=1\linewidth]{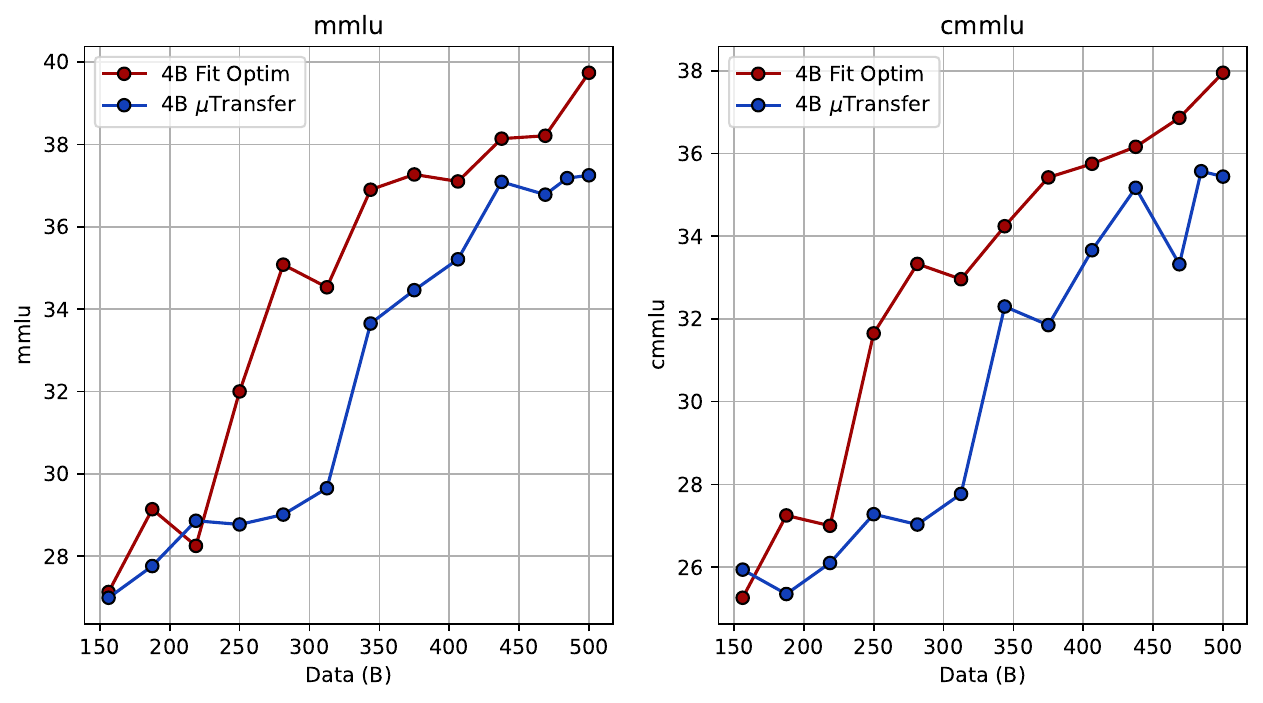}
    \caption{Downstream task performances of 4B model with global optimal LR and $\mu$P respectively.}
    \label{fig:oc_4B}
\end{figure}

\begin{figure}
    \centering
    \includegraphics[width=1\linewidth]{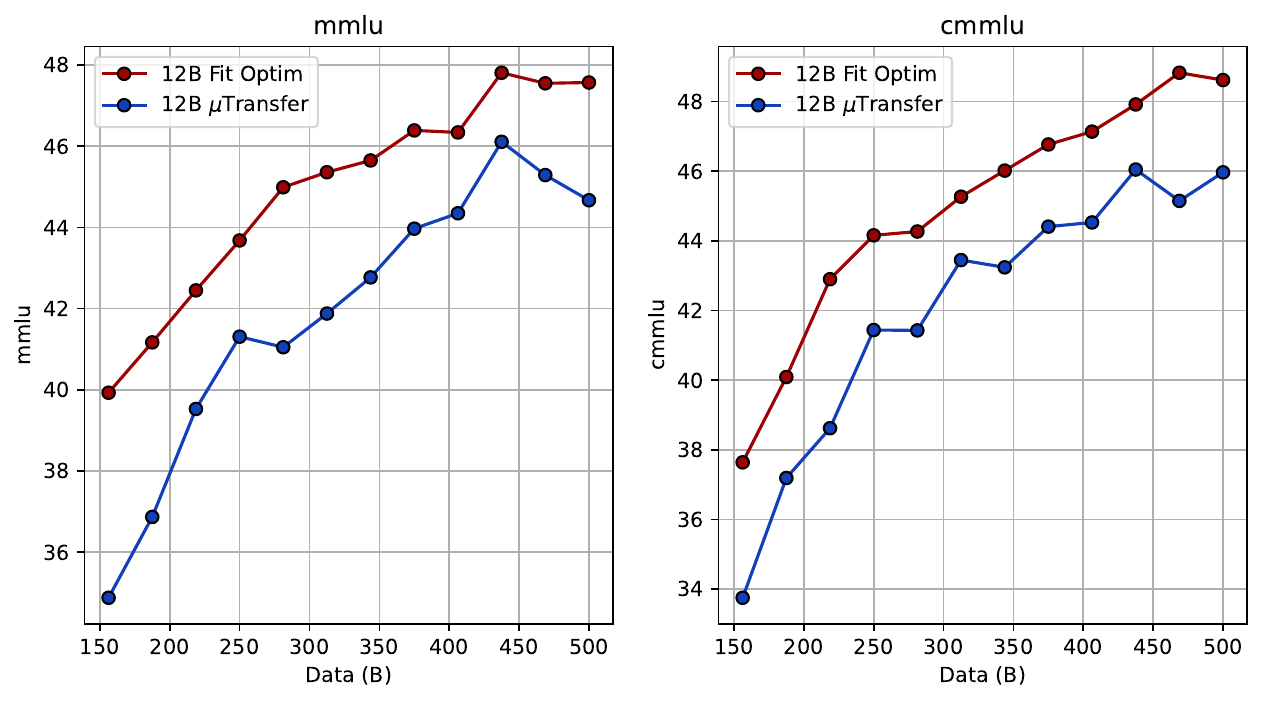}
    \caption{Downstream task performances of 12B model with global optimal LR and $\mu$P respectively.}
    \label{fig:oc_12B}
\end{figure}

First, we extrapolate the proxy model solely by increasing its width, scaling it to 4B total parameters with 530M active parameters, and conducting from-scratch pre-training on 500B tokens. To rigorously assess the pre-training quality under both paradigms, we evaluate not only the final model performance but also the downstream task results throughout the training process. The performance trends are illustrated in Figure~\ref{fig:oc_4B}. As shown, the pre-training quality achieved by the Fitting Paradigm is significantly higher than that of $\mu Transfer$.

Furthermore, we extend the predictive training scale by more than an order of magnitude, scaling the model to 12B total parameters (1.3B active) and pre-training it from scratch on 500B tokens. We similarly evaluate the intermediate progress, with the overall performance trajectories presented in Figure~\ref{fig:oc_12B}. As demonstrated in Figure~\ref{fig:oc_12B}, the model trained via the Fitting Paradigm consistently and significantly outperforms the one using $\mu Transfer$.

\section{Analyze}

\subsection{Module-Level Optimal Learning Rates}
\label{sec:module-lr}

A fundamental motivation behind $\mu$P is the hypothesis that under Standard Parametrization (SP) and a uniform global learning rate, specific modules may suffer from insufficient training, thereby failing to satisfy the regime of maximal feature learning. To investigate this, building upon the global optimal learning rate derived from our fitting paradigm in Section \ref{sec:lr-scaling-law}, we employ a greedy search strategy to conduct a fine-grained learning rate search across four distinct parameter modules: Embeddings, LM Head, Router, and Hidden parameters.We observe that fine-grained tuning of individual modules yields no significant performance improvement compared to the global optimal learning rate configuration.

\begin{figure}[t]
    \centering
    
    \subfigure[LM Head]
    {
        \label{fig:lm_head_exm}
        \includegraphics[width=0.35\linewidth]{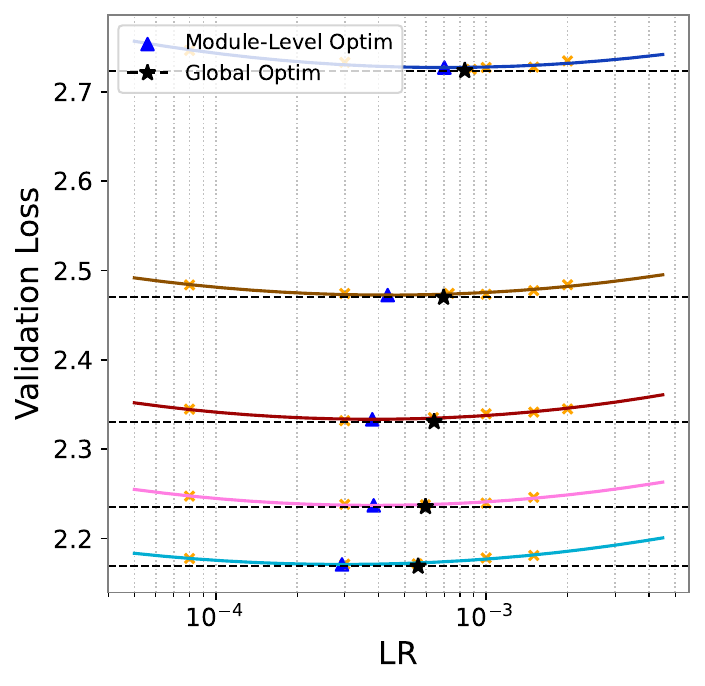}
    }
    \quad
    \subfigure[Router]
    {
        \label{fig:router_exm}
        \includegraphics[width=0.35\linewidth]{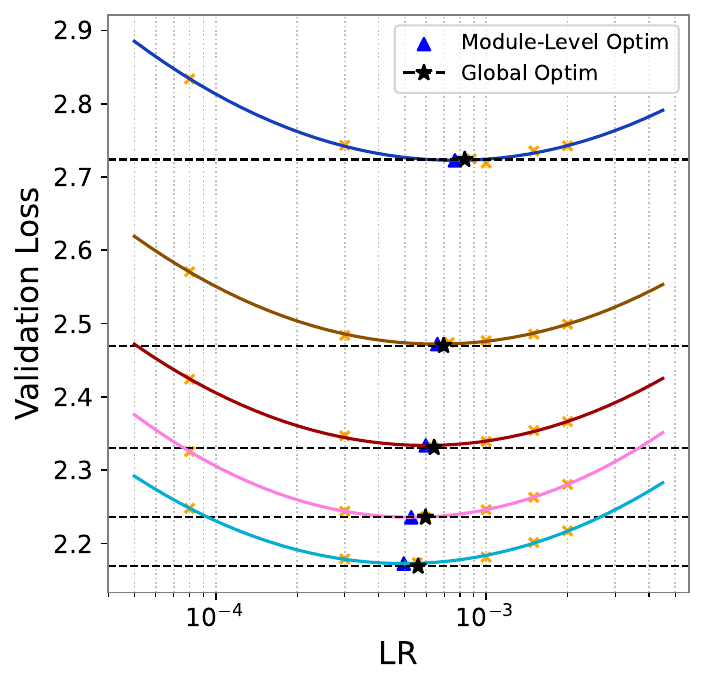}
    }

    \subfigure[Hidden]
    {
        \label{fig:hidden_exm}
        \includegraphics[width=0.35\linewidth]{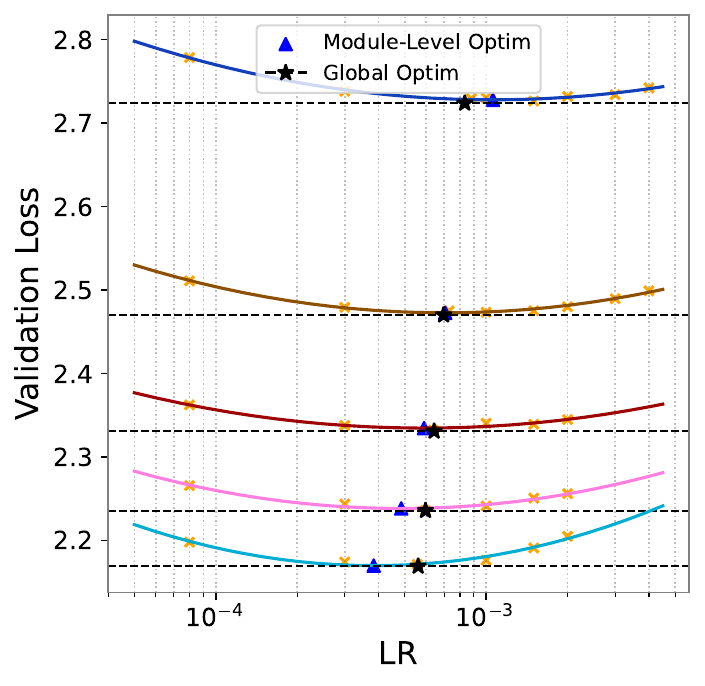}
    }
    \quad
    \subfigure[Embeddings]
    {
        \label{fig:embed_exm}
        \includegraphics[width=0.35\linewidth]{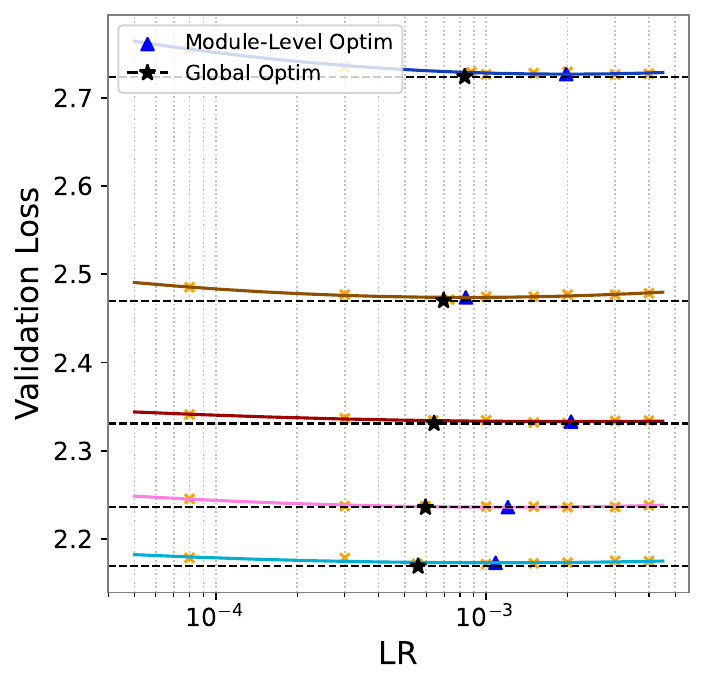}
    }
    \caption{The relationship between loss and learning rate for \textbf{(a)LM Head}, \textbf{(b)Router}, \textbf{(c)Hidden} and \textbf{(d)Embedding} parameters during the module-level learning rate search. Each curve corresponds to a model of a specific size. The dashed lines indicate the loss achieved by the corresponding model size under the global optimal learning rate setting. Triangle markers denote the optimal learning rate for the current module, while star markers represent the global optimal learning rate.}
    \label{fig:module-lr-exm}
\end{figure}

The optimal learning rates identified for specific modules align closely with the global optimum, and the minimum loss achieved through module-specific search exhibits negligible deviation from the loss achieved with global optimal learning rate from Equation \ref{eq:lr-n-d-res} (as illustrated in Figure \ref{fig:module-lr-exm}). Consequently, assigning distinct optimal learning rates to specific modules does not appear to materially enhance model performance.

\begin{figure}
    \centering
    \includegraphics[width=1\linewidth]{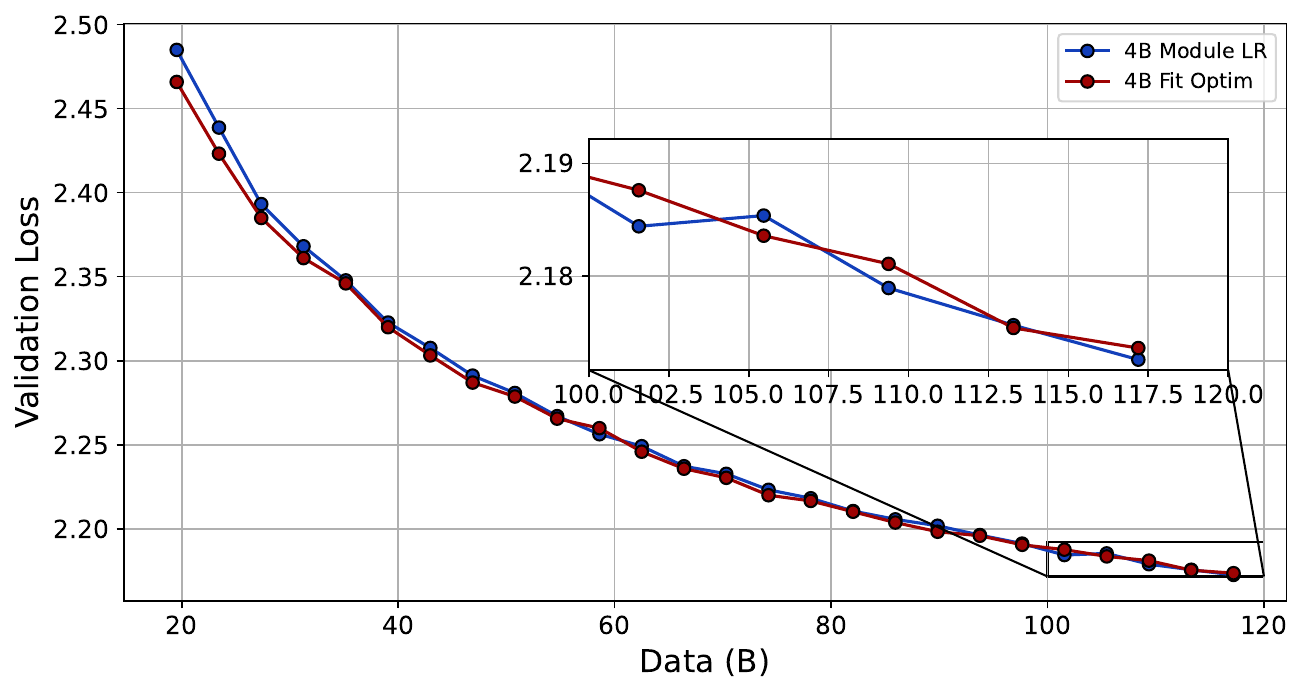}
    \caption{Performance comparison between the global optimal LR (red line) and module-wise optimal LR (blue line) on a 4B model trained for 120B tokens. The two loss curves are virtually indistinguishable during the mid-to-late stages of training ($\Delta L \le 0.01$), indicating that module-specific learning rate optimization does not yield significant performance improvements.}
    \label{fig:module_lr_res}
\end{figure}

To further validate the effect of module-level optimal LR, we trained the target 4B model for 120B tokens using both the derived module-specific optimal LR and the calculated global optimal LR. As depicted in Figure \ref{fig:module_lr_res}, the comparison of the validation losses reveals that the loss curves for both settings are virtually indistinguishable. See Table \ref{tab:module-lr-settings} and Appendix \ref{sec:appendix-module-lr} for detailed settings.

\subsection{A Closer Look at Feature Learning}
\label{sec:feature-learning}

In the previous subsection, we observed that fine-grained learning rate tuning across distinct model modules yielded no substantial performance gains, indicating that a global learning rate configuration does not induce training imbalances among components. In this subsection, we further investigate the feature learning dynamics of these modules by analyzing the optimization trajectory, specifically monitoring the evolution of parameter update magnitudes throughout the training process.

\begin{figure}
    \centering
    \vspace{-0.35cm}
    \subfigtopskip=2pt
    \subfigbottomskip=2pt
    \subfigcapskip=-5pt
    \subfigure[4B Model]
    {
        \label{fig:update_norm_4B}
        \includegraphics[width=1\linewidth]{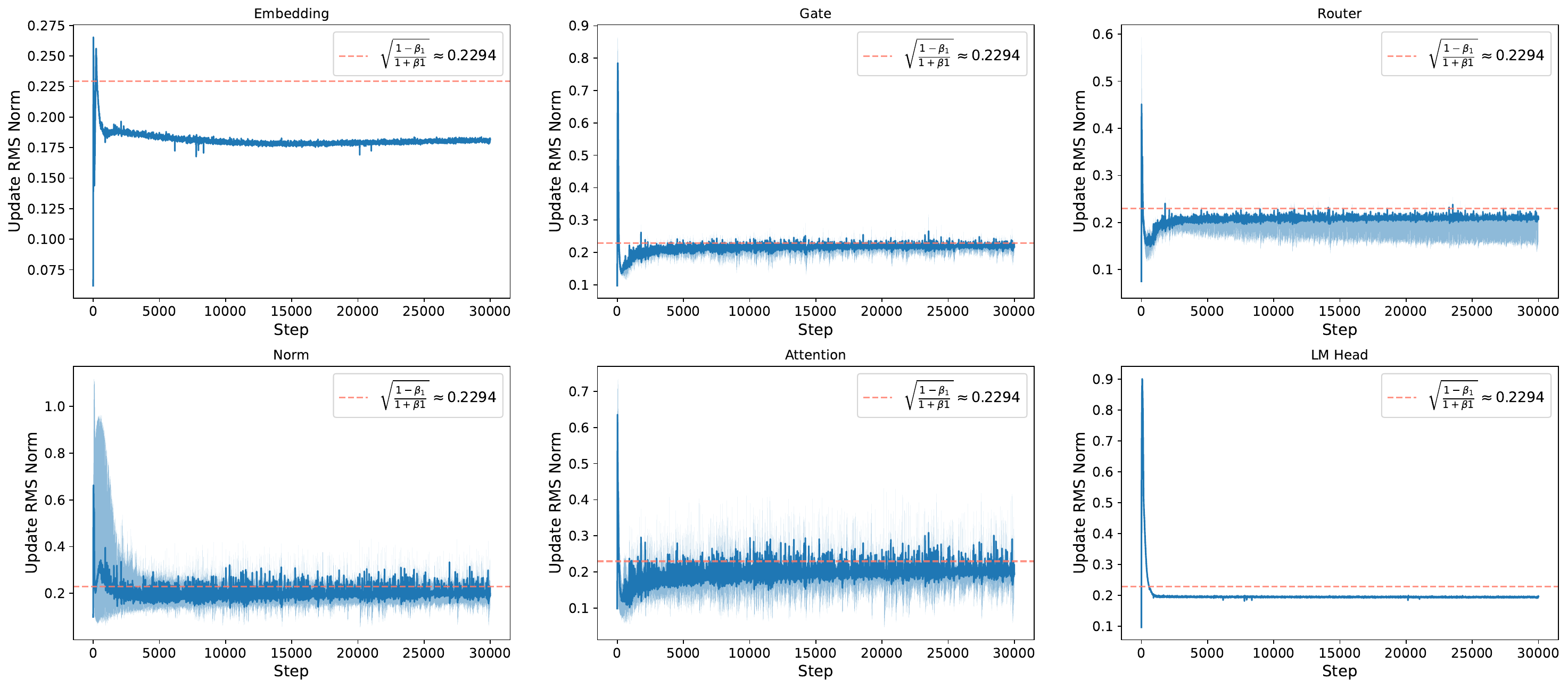}
    }

    \subfigure[500M Model without QK-Norm]
    {
        \label{fig:update_norm_500M_nonorm}
        \includegraphics[width=1\linewidth]{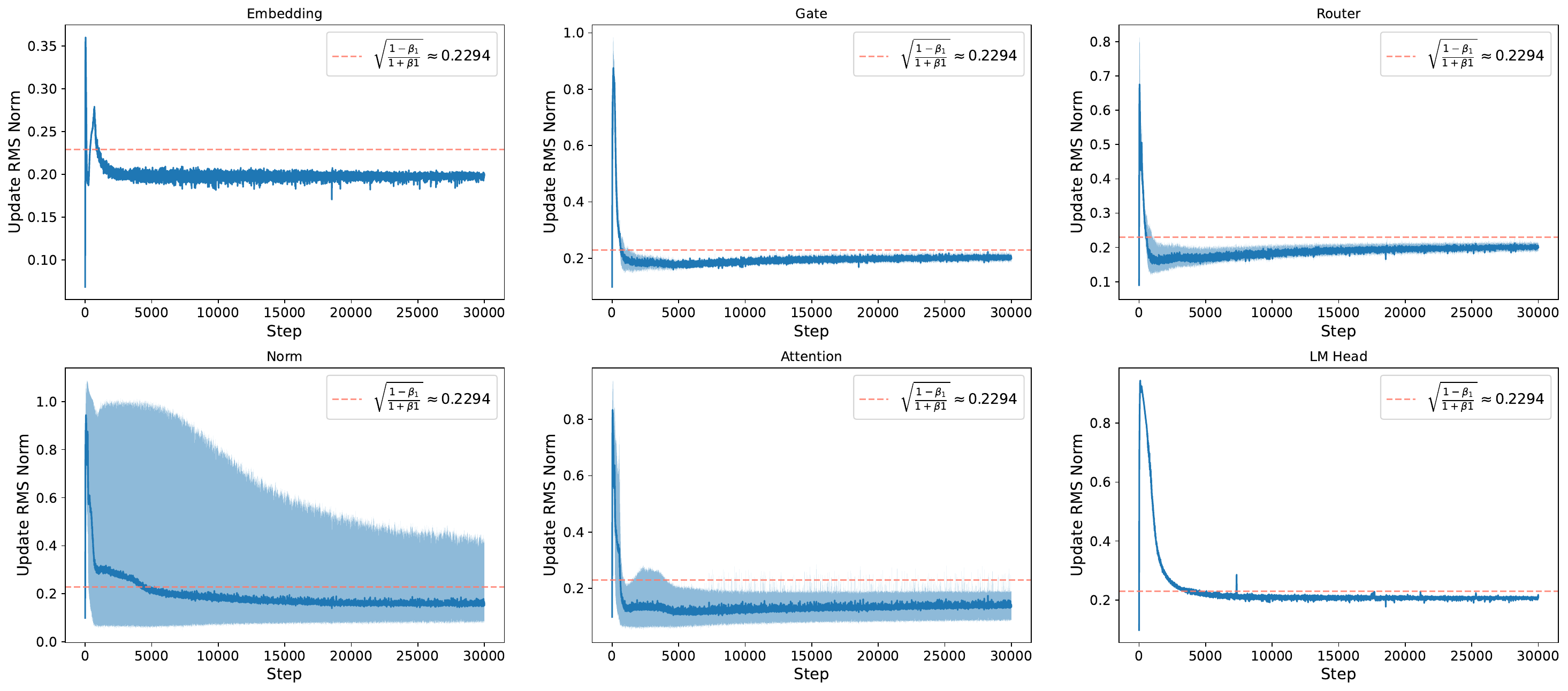}
    }

    \subfigure[4B $\mu$P Model]
    {
        \label{fig:update_norm_mup}
        \includegraphics[width=1\linewidth]{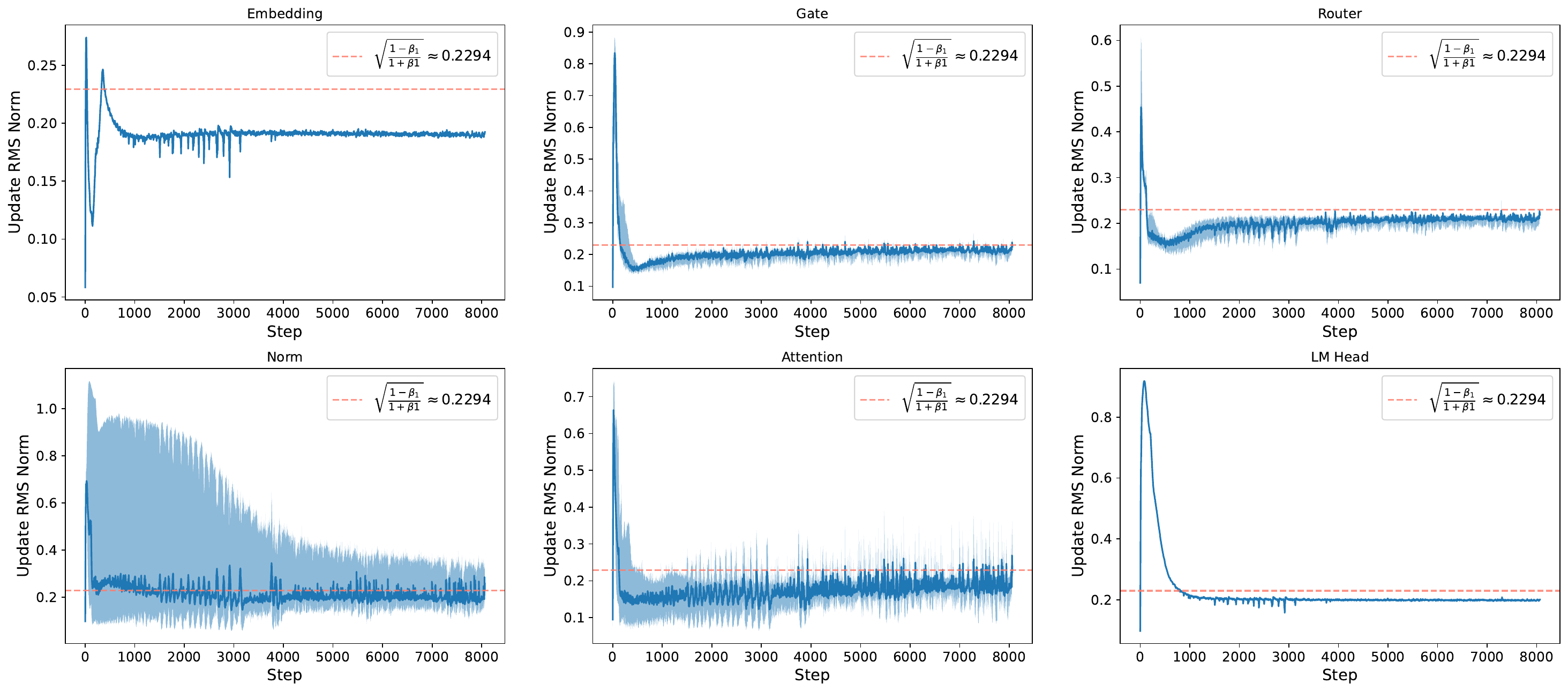}
    }
    \caption{Update RMS norm during the training process of: (a) 4B model (b) 500M model with QK-Norm removed (c) 4B model with $\mu$P. Update RMS norm maintains approximately 0.2 in all the models we observed.}
    \label{fig:update_norm}
\end{figure}

As illustrated in the Figure\ref{fig:update_norm}, training with the AdamW optimizer results in parameter update magnitudes that remain stable over extended periods and exhibit relative uniformity across layers. The update magnitudes consistently approximate 0.2, a finding consistent with recent theoretical studies \cite{liu2025muonscalablellmtraining,rotationalequilibrium}. This evidence further corroborates that distinct modules maintain comparable feature learning capabilities at any given stage of training, thereby negating the necessity for module-specific learning rates to balance feature learning efficiency.

\subsection{Does Standard Parametrization Scale?}
\label{sec:logits}

Training stability is widely regarded as another distinct advantage of $\mu$P. \citet{mup} argues that under standard parametrization, the internal training states of certain modules tend to "blow up" as model scale increases, thus the adjustment of learning rates on different modules is necessary.

\begin{figure}
    \centering
    \vspace{-0.35cm}
    \subfigtopskip=2pt
    \subfigbottomskip=2pt
    \subfigcapskip=-5pt
    \subfigure[4B Model]
    {
        \label{fig:logits_global_size}
        \includegraphics[width=1\linewidth]{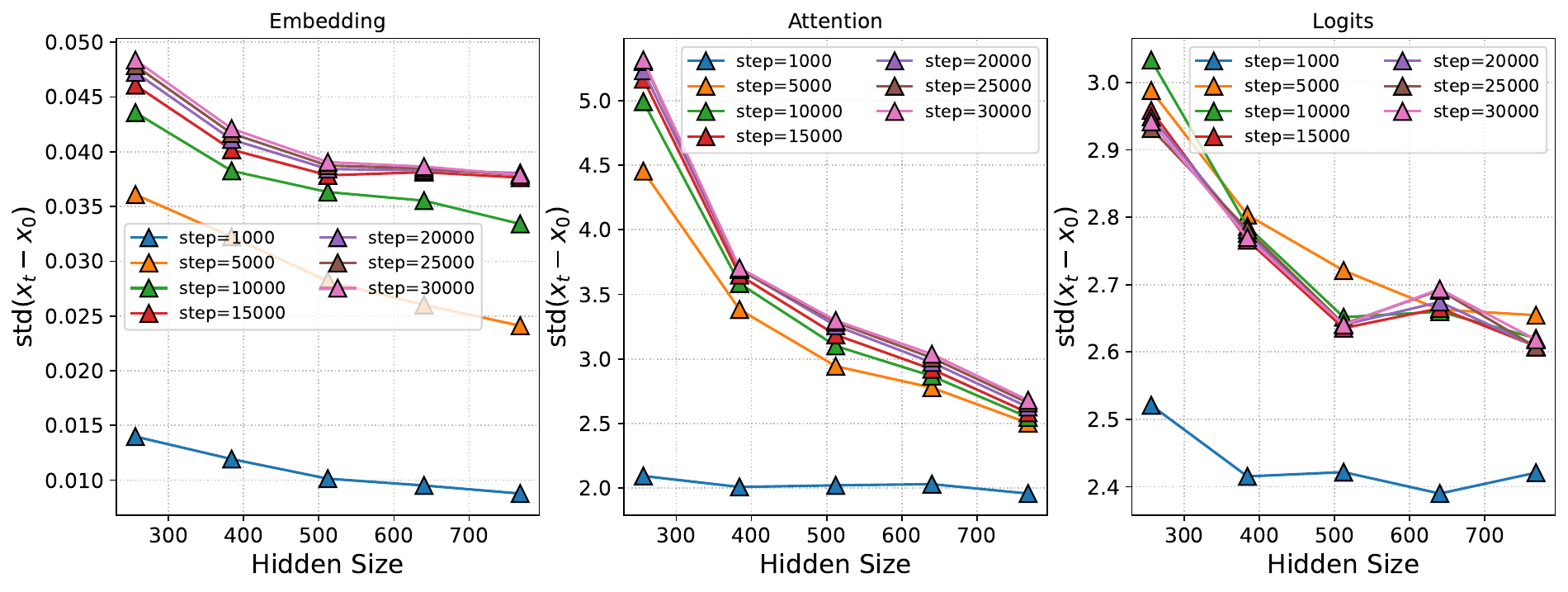}
    }

    \subfigure[4B $\mu$P Model]
    {
        \label{fig:logits_mup}
        \includegraphics[width=1\linewidth]{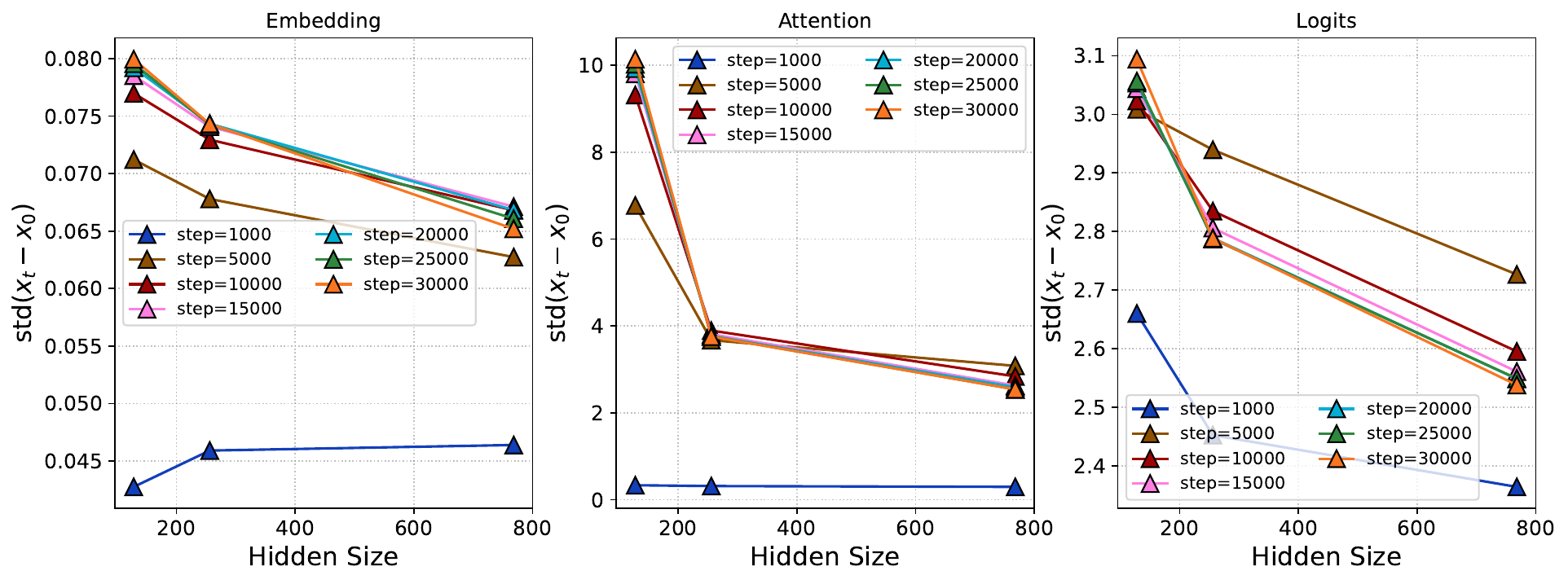}
    }

    \caption{Variation of word embeddings, attention logits, and logits compared to initial states at certain training steps as width increases. With reference to \citet{mup}, we plot the standard deviation of the coordinates of $x_t - x_0$, $x\in \{word\ embeddings, attention\ logits, logits\}$. In our experiments, logits and attention logits of models with standard parametrization do not exhibit the "blow-up" tendency.}
    \label{fig:logits}
\end{figure}

\begin{figure}
    \centering
    \vspace{-0.35cm}
    \subfigtopskip=2pt
    \subfigbottomskip=2pt
    \subfigcapskip=-5pt
    \subfigure[4B Model]
    {
        \label{fig:logits_global_size_attn}
        \includegraphics[width=0.43\linewidth]{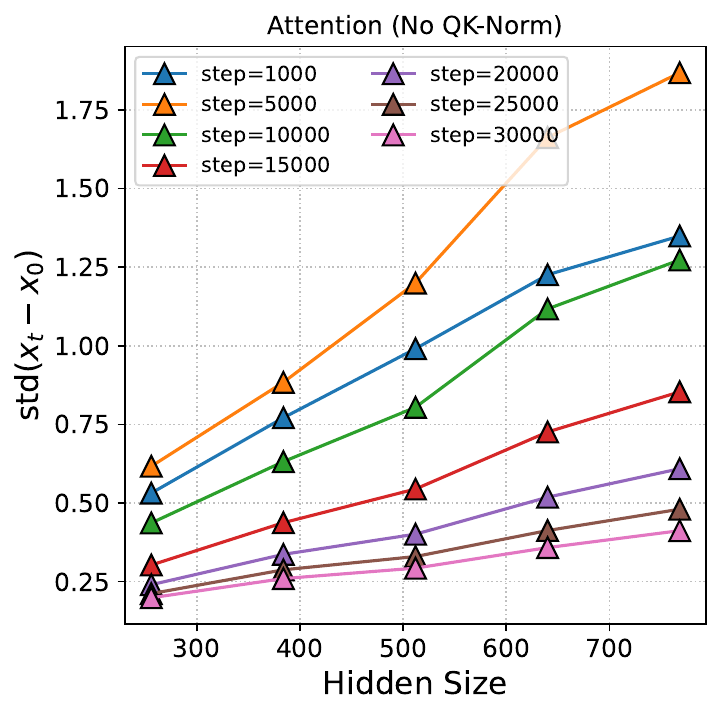}
    }
    \subfigure[4B $\mu$P Model]
    {
        \label{fig:logits_mup_attn}
        \includegraphics[width=0.43\linewidth]{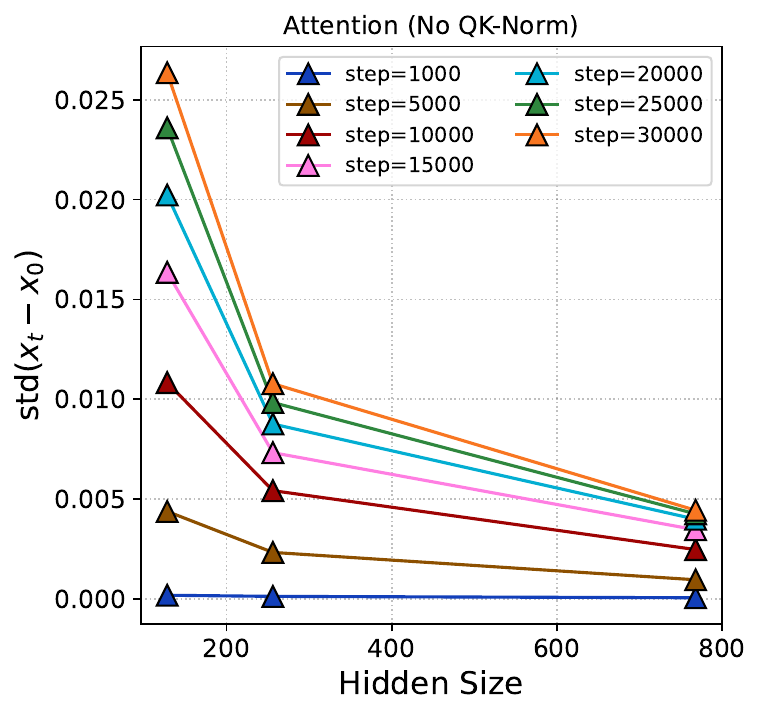}
    }
    \caption{Variation of attention logits at certain training steps as width increases. We ignored QK-Norm parameters when compute attention logits. Attention logits started to blow up with width in SP model. }
    \label{fig:logits_attn}
\end{figure}

We replicated the methodology of $\mu$Transfer to analyze the model derived from our experiments. Contrary to expectations, under standard parametrization, the internal states of our model did not exhibit blow up(Figure \ref{fig:logits_global_size}); rather, they displayed trends remarkably similar to those of models initialized via $\mu$P. To investigate further, we conducted an ablation where the QK-Norm modules were removed during the computation of attention logits (Figure \ref{fig:logits_attn}). Under this condition, we successfully reproduced the instability trends described in $\mu$Transfer. Consequently, we posit that recent advancements in model architecture—such as the incorporation of QK-Norm—have rendered layer-wise training more balanced and significantly enhanced robustness to hyperparameter variations.

\subsection{Impact of Data Size on Training Stability}
\label{sec:logits_step}

Existing research on training stability, most notably the work on $\mu$P, has predominantly focused on model scale while neglecting the influence of training data size. Employing the analytical framework established in Section \ref{sec:logits}, we investigate the evolution of the model's internal states as the amount of training data increases under standard parametrization.

\begin{figure}
    \centering
    \includegraphics[width=1\linewidth]{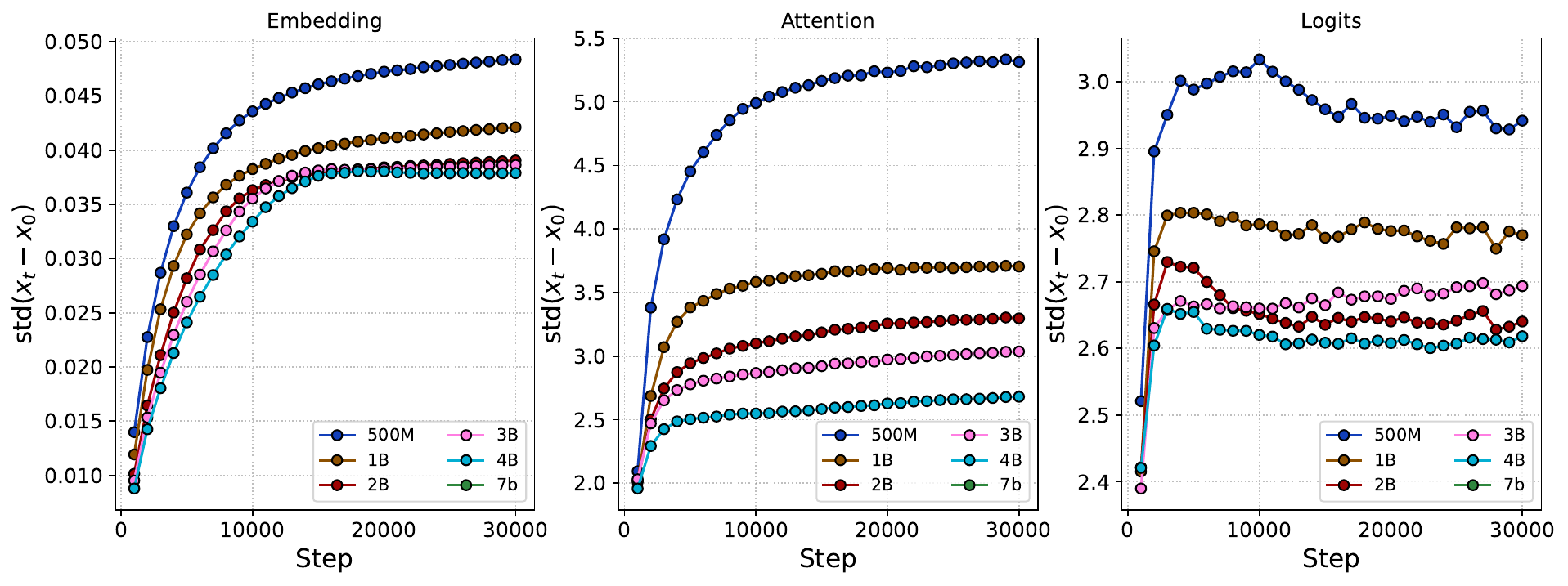}
    \caption{Variation of word embeddings, attention logits, and logits compared to initial states as training proceeded.}
    \label{fig:logits_step}
\end{figure}

As illustrated in the Figure \ref{fig:logits_step}, while the model's internal states eventually converge to a relatively stable regime as the amount of training data increases, the internal variations are significantly more drastic with respect to data scaling than to model scaling. This phenomenon is particularly evident in the attention logits. This observation offers a potential explanation for the scaling coefficients in Equation \ref{eq:lr-n-d-res}, where the exponent for model parameter count $N$ ($-0.22$) is algebraically greater than that for data volume $D$ ($-0.35$). As the size of training data expands, the magnitude of parameter updates across different modules exhibits a more pronounced increase; consequently, the optimal learning rate requires more substantial adjustment to maintain training stability.

\subsection{Decay Training}
\label{sec:decay}
A key characteristic of WSD schedule is the utilization of higher-quality training data during the decay phase after the constant-learning-rate stable phase to maximize the model's feature learning.

Building upon the experiments described in Section \ref{sec:results} we extended the training of both model variants after the end of the stable phase using a distinct corpus of high-quality data. We annealed the learning rate to 10\% of its value during the stable phase and continued training for an additional 100B tokens. We then evaluated the downstream task performance of the models that had completed the full WSD training.

\begin{figure}
    \centering
    \includegraphics[width=1\linewidth]{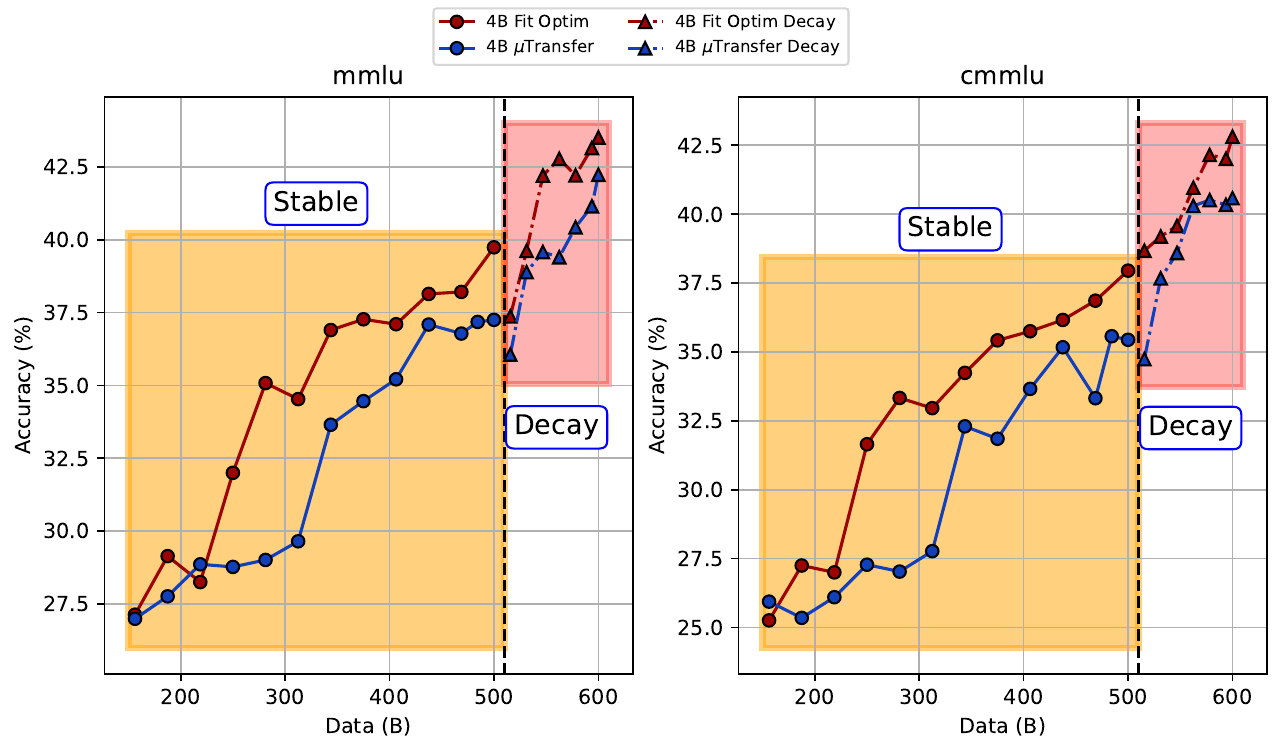}
    \caption{Downstream task performances of 4B model after the end of decay phase. The left area depicts the stable phase of training, while the right area corresponds to the decay phase.}
    \label{fig:oc_4B_decay}
\end{figure}

As shown in Figure \ref{fig:oc_4B_decay}, the model trained with the global optimal learning rate outperformed the model derived from $\mu$ Transfer on both the MMLU and CMMLU benchmarks, achieving accuracy improvements of 1.28\% ($42.23\% \rightarrow 43.51\%$) and 2.23\% ($40.58\% \rightarrow 42.81\%$), respectively. These results demonstrate that the global optimal learning rate yields superior performance in realistic pre-training scenarios.

\section{Conclusion}
This paper systematically establishes two fundamental research paradigms—Fitting and Transfer—to address the critical challenge of learning rate configuration in large-scale pre-training. At the methodological level, we introduce scaling Laws to reduce the complexity of the Fitting Paradigm, and provide a comprehensive extension of $\mu$Transfer across model architectures, depths, weight decay, and token horizons. Through extensive experimentation, we challenge the scalability of the widely adopted $\mu$Transfer in large-scale pre-training scenarios and provide an in-depth analysis of the underlying mechanisms that limit the performance of module-wise parameter tuning at scale. This research offers both systematic practical guidance and a novel theoretical perspective for optimizing industrial-level pre-training.

\section*{Limitations}
To inform and inspire future research, we summarize the limitations of our work as follows:\\

Learning Rate Schedules: This study focuses on large-scale pre-training, where the Warm-Stable-Decay (WSD) scheduler is currently the industry standard. Consequently, our analysis is centered on this specific schedule and does not explore the dynamics of other learning rate schedulers.\\

Model Architectures: Given that the Mixture of Experts (MoE) architecture has become the foundational backbone for modern large-scale pre-training, it served as the primary subject of our investigation. The generalizability of our findings to Dense architectures remains to be verified in future work.\\

Extrapolation Limits: Due to computational resource constraints, 
this study did not investigate the ultimate extrapolation boundaries (i.e., the maximum scale at which these predictions remain accurate) for both the Fitting and Transfer paradigms.
\section*{Use of AI Assistants}

We primarily use AI assistants to improve and enrich our writing.

\clearpage
\bibliographystyle{plainnat}
\bibliography{custom}


\clearpage
\appendix

\section{Appendix}
\label{sec:appendix}

\subsection{Model Architectures and LR Settings}
\label{sec:appendix-models}

Table \ref{tab:models}, \ref{tab:mup-models} and \ref{tab:module-lr-settings} shows the detailed parameters of models' architectures and learning rate settings.

\begin{table*}[htbp]
    \centering
    \caption{Overview of Qwen3-MoE Model Architectures and Hyperparameters in Section \ref{sec:lr-scaling-law}}
    \label{tab:models}
    
    \resizebox{\textwidth}{!}{%
        \begin{tabular}{lcccccccc}
            \toprule
            \textbf{Models} & 
            \textbf{\shortstack{Total\\Params}} & 
            \textbf{\shortstack{Activate\\Params}} & 
            \textbf{\shortstack{Hidden\\Size}} & 
            \textbf{\shortstack{Num\\Layers}} & 
            \textbf{\shortstack{Attn\\Heads}} & 
            \textbf{\shortstack{KV\\Heads}} & 
            \textbf{\shortstack{Interm.\\Size}} & 
            \textbf{Learning Rate} \\
            \midrule
            
            \multicolumn{9}{c}{\textit{\textbf{Training Set}}} \\
            \midrule
            Qwen3-MoE-0.5B-A0.1B & 550M & 100M & 256  & 3  & 32 & 4 & 768 & 8e-5, 1e-4, 3e-4, 5e-4, 8e-4, 1.5e-3, 2e-3 \\
            Qwen3-MoE-1B-A0.2B   & 1B   & 190M & 384  & 9  & 32 & 4 & 768 & 8e-5, 1e-4, 3e-4, 5e-4, 8e-4, 1.5e-3, 2e-3 \\
            Qwen3-MoE-2B-A0.3B   & 2B   & 280M & 512  & 12 & 32 & 4 & 768 & 8e-5, 1e-4, 3e-4, 5e-4, 8e-4, 1.5e-3 \\
            Qwen3-MoE-3B-A0.4B   & 3B   & 400M & 640  & 15 & 32 & 4 & 768 & 8e-5, 1e-4, 3e-4, 5e-4, 8e-4, 1.5e-3 \\
            \midrule
            
            \multicolumn{9}{c}{\textit{\textbf{Test Set}}} \\
            \midrule
            Qwen3-MoE-4B-A0.5B   & 4B   & 530M & 768  & 18 & 32 & 4 & 768 & 8e-5, 3e-4, 5e-4, 8e-4, 1.5e-3 \\
            Qwen3-MoE-12B-A1.3B  & 12B  & 1.3B & 1280 & 30 & 32 & 4 & 768 & - \\
            \bottomrule
        \end{tabular}%
    }
    \vspace{1ex}
\end{table*}

\begin{table*}[htbp]
    \centering
    \caption{Overview of Qwen3-MoE Model Architectures and Hyperparameters in Section \ref{sec:mup}}
    \label{tab:mup-models}
    
    \resizebox{\textwidth}{!}{%
        \begin{tabular}{lccccccccc}
            \toprule
            \textbf{Models} & 
            \textbf{\shortstack{Total\\Params}} & 
            \textbf{\shortstack{Activate\\Params}} & 
            \textbf{\shortstack{Hidden\\Size}} & 
            \textbf{\shortstack{Num\\Layers}} & 
            \textbf{\shortstack{Attn\\Heads}} & 
            \textbf{\shortstack{KV\\Heads}} & 
            \textbf{\shortstack{Interm.\\Size}} & 
            \textbf{Learning Rate} & 
            \textbf{std} \\
            \midrule
            
            Qwen3-MoE-2B-A0.3B-proxy & 2B & 290M & 512 & 18 & 32 & 4 & 512 & 
            \shortstack[l]{8e-5, 1e-4, 3e-4, 5e-4,\\ 1e-3, 2e-3} & 
            0.01, 0.015, 0.02, 0.03, 0.04 \\
            
            Qwen3-MoE-4B-A0.5B & 4B & 530M & 768 & 18 & 32 & 4 & 768 & - & - \\
            
            \addlinespace 
            
            Qwen3-MoE-2B-A0.3B-proxy & 2B & 290M & 640 & 18 & 32 & 4 & 384 & 
            \shortstack[l]{1e-4, 3e-4, 5e-4, 1e-3,\\ 2e-3} & 
            \shortstack[l]{0.0005, 0.001, 0.002,\\ 0.005, 0.01, 0.02} \\
            
            Qwen3-MoE-12B-A1.3B & 12B & 1.3B & 1280 & 30 & 32 & 4 & 768 & - & - \\
            
            \bottomrule
        \end{tabular}%
    }
    \vspace{1ex}
\end{table*}

\begin{table*}[htbp]
    \centering
    \caption{Overview of Qwen3-MoE Model Architectures and Hyperparameters in Section \ref{sec:module-lr}}
    \label{tab:module-lr-settings}
    
    \resizebox{\textwidth}{!}{%
        \begin{tabular}{lcccccccc}
            \toprule
            \textbf{Models} & 
            \textbf{\shortstack{Total\\Params}} & 
            \textbf{\shortstack{Activate\\Params}} & 
            \textbf{\shortstack{Hidden\\Size}} & 
            \textbf{\shortstack{Num\\Layers}} & 
            \textbf{\shortstack{Attn\\Heads}} & 
            \textbf{\shortstack{KV\\Heads}} & 
            \textbf{\shortstack{Interm.\\Size}} & 
            \textbf{Learning Rate} \\
            \midrule
            
            \multicolumn{9}{c}{\textit{\textbf{Training Set}}} \\
            \midrule
            Qwen3-MoE-0.5B-A0.1B & 550M & 100M & 256  & 3  & 32 & 4 & 768 & 8e-5, 3e-4, 8.75e-4, 1e-3, 1.5e-3, 2e-3, 3e-3, 4e-3 \\
            Qwen3-MoE-1B-A0.2B   & 1B   & 190M & 384  & 9  & 32 & 4 & 768 & 8e-5, 3e-4, 7.24e-4, 1e-3, 1.5e-3, 2e-3, 3e-3, 4e-3 \\
            Qwen3-MoE-2B-A0.3B   & 2B   & 280M & 512  & 12 & 32 & 4 & 768 & 8e-5, 3e-4, 6.36e-4, 1e-3, 1.5e-3, 2e-3, 3e-3, 4e-3 \\
            Qwen3-MoE-3B-A0.4B   & 3B   & 400M & 640  & 15 & 32 & 4 & 768 & 8e-5, 3e-4, 5.90e-4, 1e-3, 1.5e-3, 2e-3, 3e-3, 4e-3 \\
            Qwen3-MoE-4B-A0.5B   & 4B   & 530M & 768  & 18 & 32 & 4 & 768 & 8e-5, 3e-4, 5.55e-4, 1e-3, 1.5e-3, 2e-3, 3e-3, 4e-3 \\
            \bottomrule
        \end{tabular}%
    }
    \vspace{1ex}
\end{table*}

\subsection{Extrapolation of L(D)}
\label{sec:appendix-ld}

\begin{figure}[t]
    \centering
    \includegraphics[width=1\linewidth]{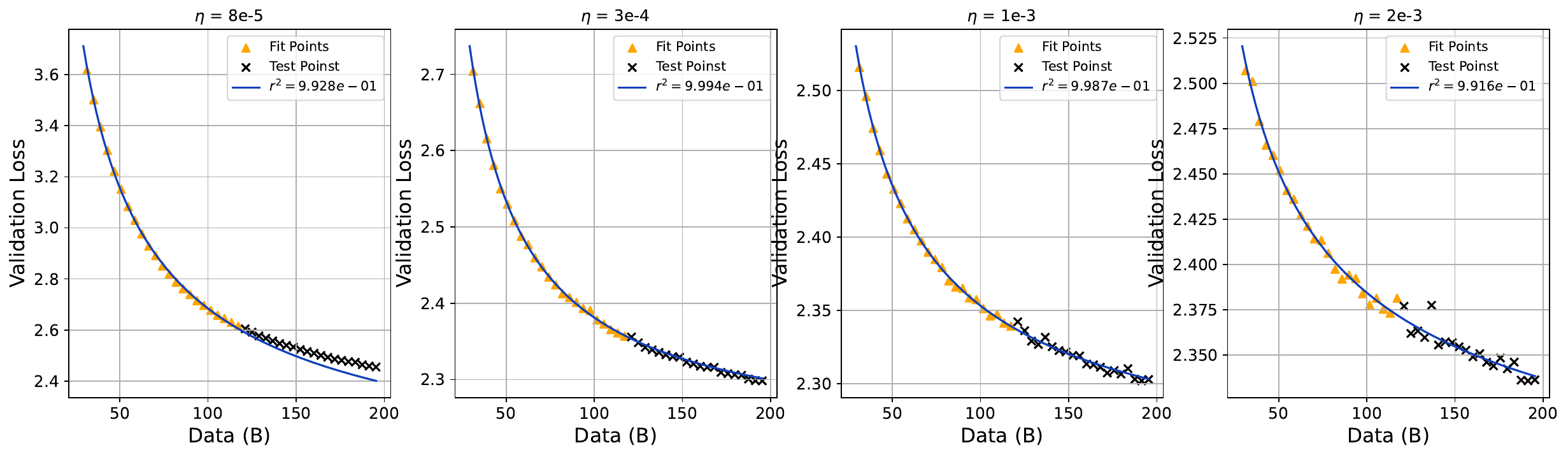}
    \caption{The precision of Equation \ref{eq:ld}.}
    \label{fig:ld_mup_v2}
\end{figure}

In the experiments presented in our works, $L(D) = L_0 + A \cdot D^{-\gamma}$(Equation \ref{eq:ld}) is repeatedly employed for curve fitting and data augment. To validate the effectiveness of this approach, we continue training the 2B proxy model of $\mu$Transfer from 120B to approximately 200B tokens (50,000 steps). Points corresponding to $D \le 120B$ are used as the fitting set, while the remaining data serve as the test set. The fitted curve is illustrated in Figure \ref{fig:ld_mup_v2}.

Among the four settings, the extrapolation accuracy is notably poorer under the learning rate of 8e-5, while the predicted values for the other learning rates at 200B tokens align closely with the ground truth. We attribute this discrepancy to the fact that the learning rate of 8e-5 is excessively small for the current model, causing the model state to evolve too gradually as data volume increases. By 120B tokens, the model has yet to exhibit a clear trend toward convergence. Consequently, the curve fitted by Equation \ref{eq:ld} declines sharply rather than gradually flattening, leading to a misinterpretation of the model's future trajectory. In contrast, the other tested learning rates are relatively larger and closer to the model's optimal learning rate, enabling the validation loss curve to enter the convergence phase more rapidly and thus yielding more accurate predictions from the L(D) curve.

\begin{figure}[t]
    \centering
    \includegraphics[width=1\linewidth]{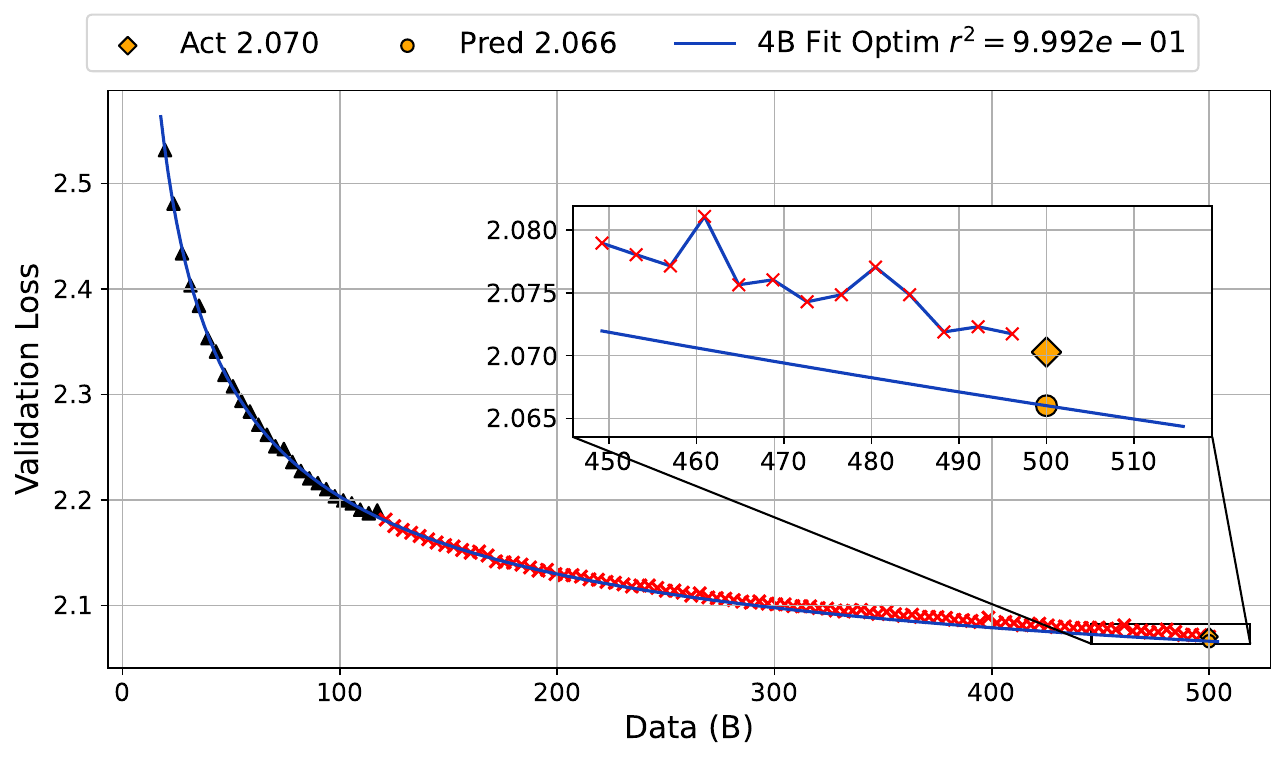}
    \caption{The precision of Equation \ref{eq:ld} on 4B Fit Optim model.}
    \label{fig:ld_exp}
\end{figure}

For experiments conducted in Section \ref{sec:results}, we also conduct the same experiment on 4B Model, fitting with points that corresponding to $D \le 120B$. Figure \ref{fig:ld_exp} indicates that despite only approximately one-quarter of the data is used for fitting, the predicted value at $D = 500B(2.066)$ exhibits a negligible discrepancy from the actual value $(2.070)$. Therefore, we consider the method of data extrapolation via Equation \ref{eq:ld} to be reasonable within the data range discussed in this paper.

\subsection{Details of Fitting Experiments}

\subsubsection{Global Optimal Learning Rate}

\label{sec:appendix-global-lr}

This subsection contains the whole fitting process of Section \ref{sec:lr-scaling-law}.

\begin{figure*}[t]
    \centering
    \includegraphics[width=1\linewidth]{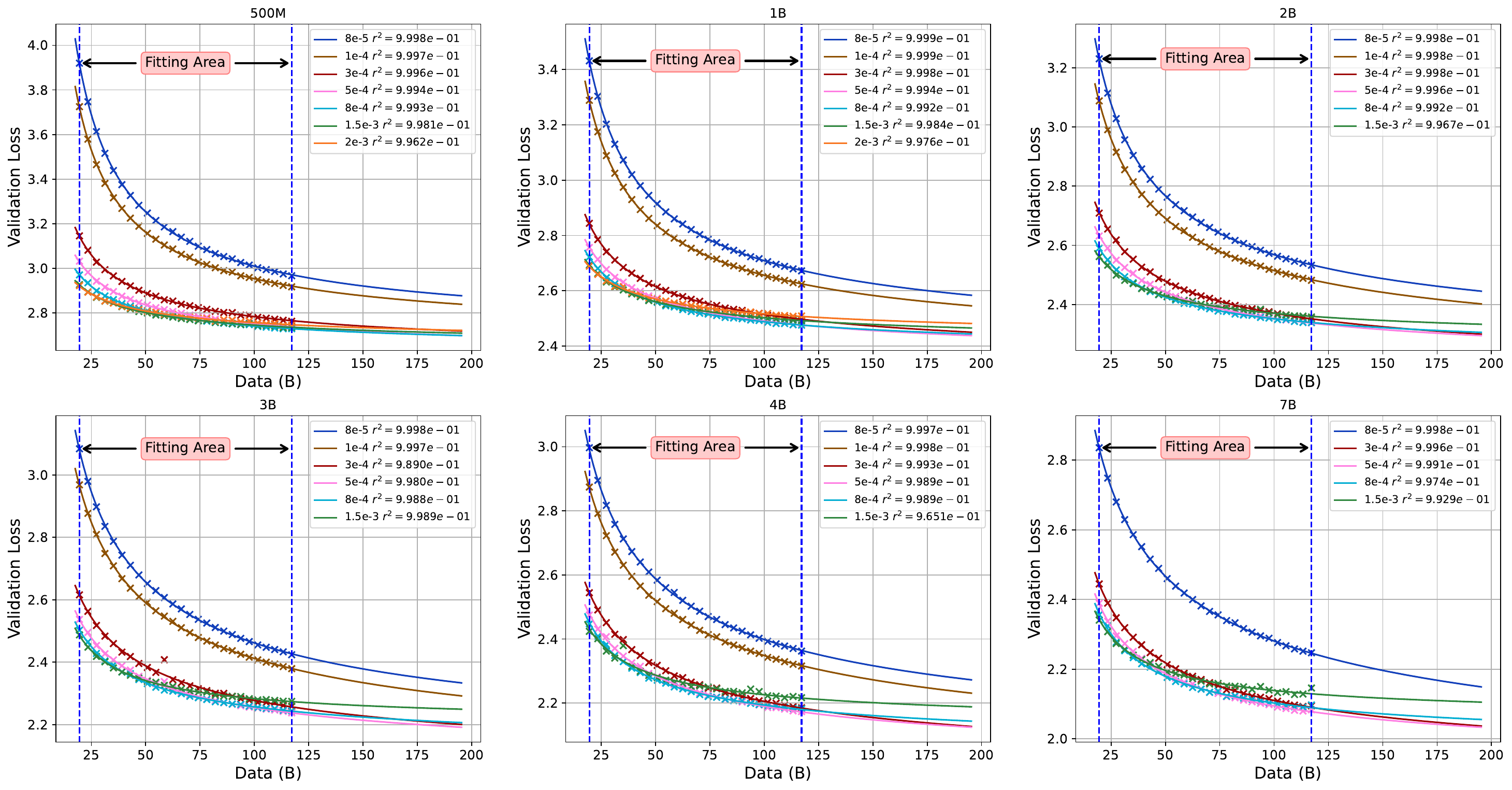}
    \caption{Results of fitting via $L(D) = L_0 + A \cdot D^{-\gamma}$ for each group of experiments.}
    \label{fig:ld_size}
\end{figure*}

The validation loss curves for various models under different learning rates, derived from the global optimal learning rate search experiments, are illustrated in Figure \ref{fig:ld_size}. We utilize the smoothed data via Equation \ref{eq:ld} as the input for subsequent fitting stages. Furthermore, we employ this equation to extrapolate the validation loss for each model at a training volume of 200B tokens across distinct learning rates, thereby augmenting the dataset available for fitting.

\begin{figure*}[t]
    \centering
    \includegraphics[width=1\linewidth]{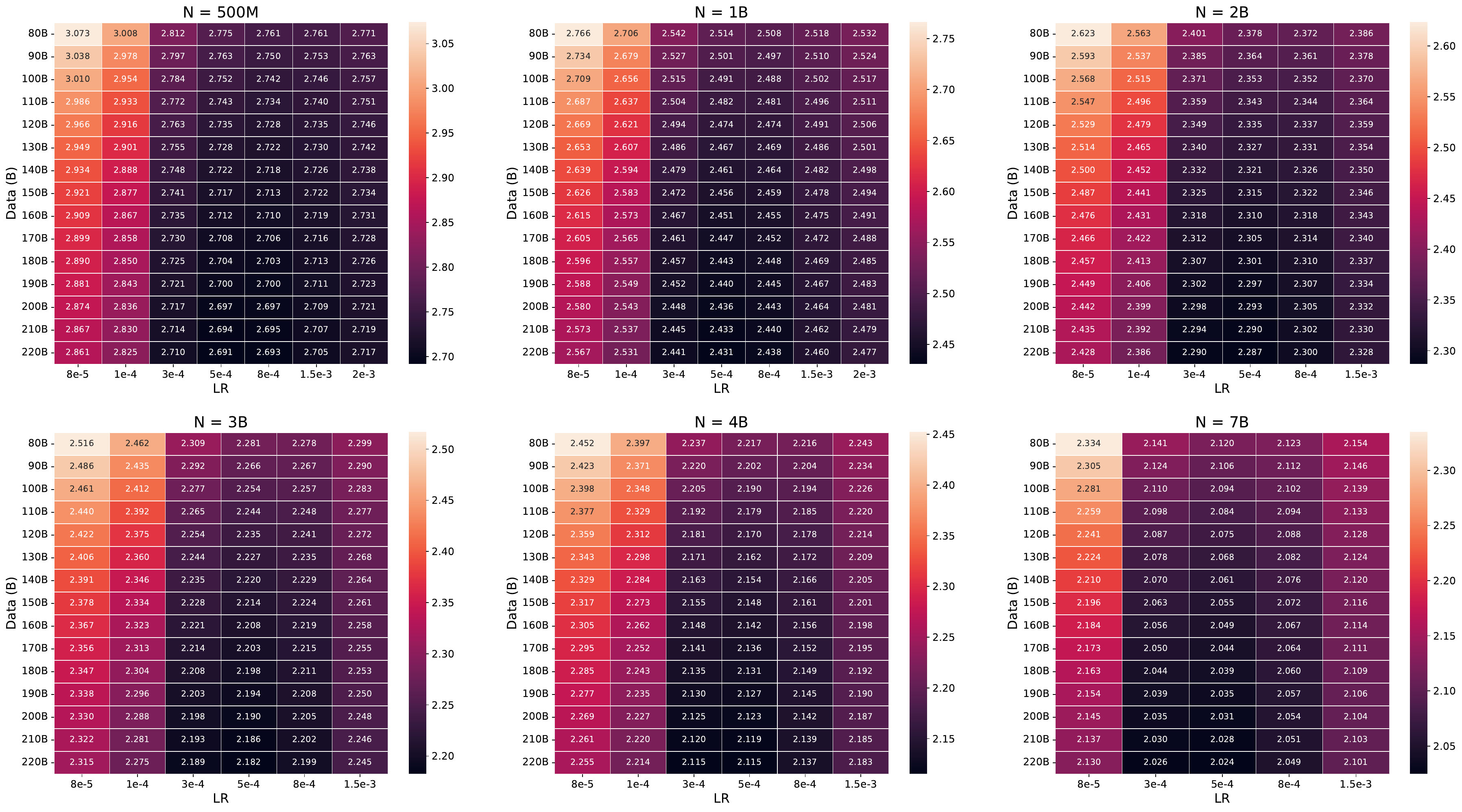}
    \caption{Relationship among loss, learning rate, and training data size of various model.}
    \label{fig:loss_lr_d_heatmap}
\end{figure*}

\begin{figure*}[t]
    \centering
    \includegraphics[width=1\linewidth]{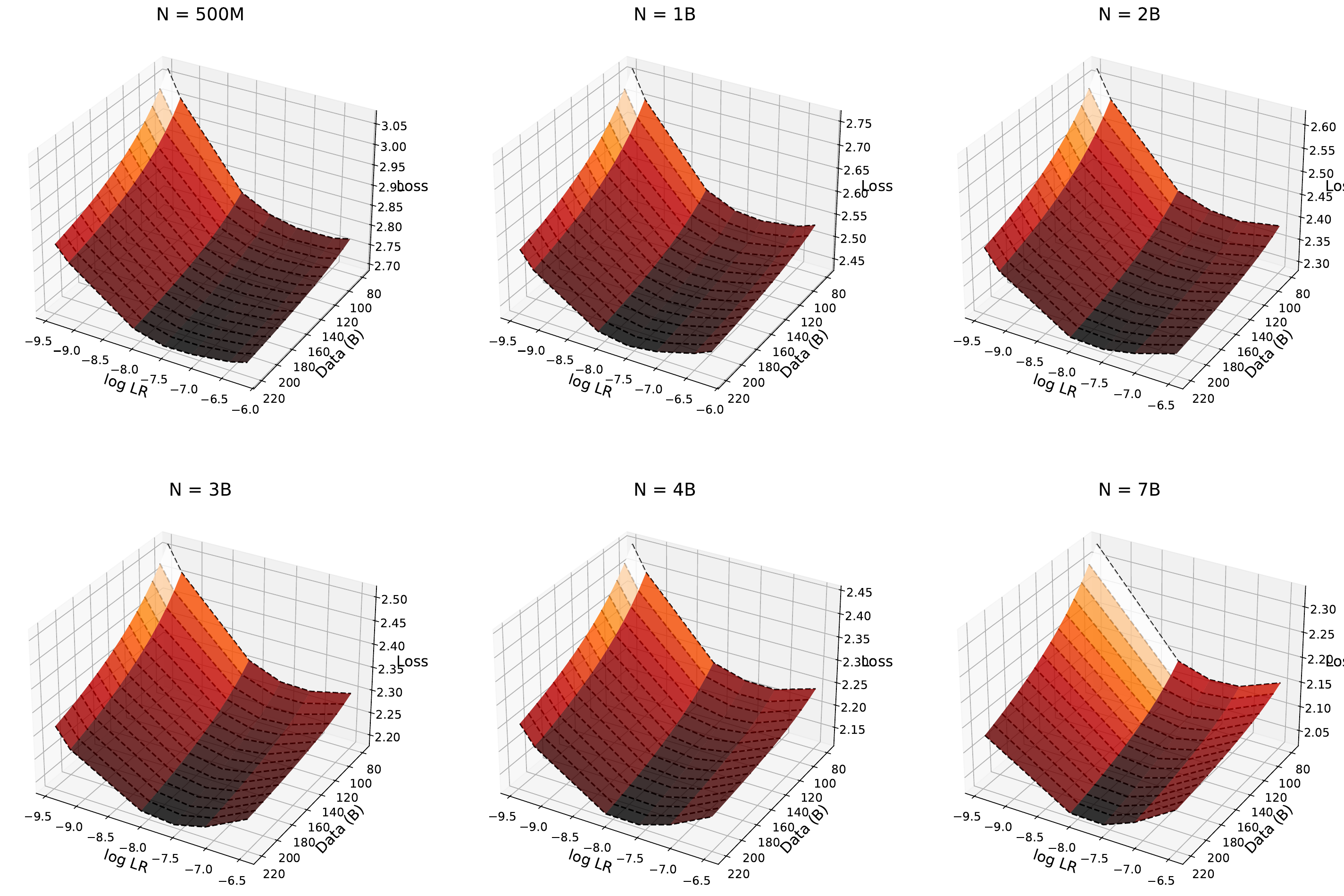}
    \caption{3D visualization of \ref{fig:loss_lr_d_heatmap}}.
    \label{fig:loss_lr_d_3d}
\end{figure*}

\begin{figure*}[t]
    \centering
    \includegraphics[width=1\linewidth]{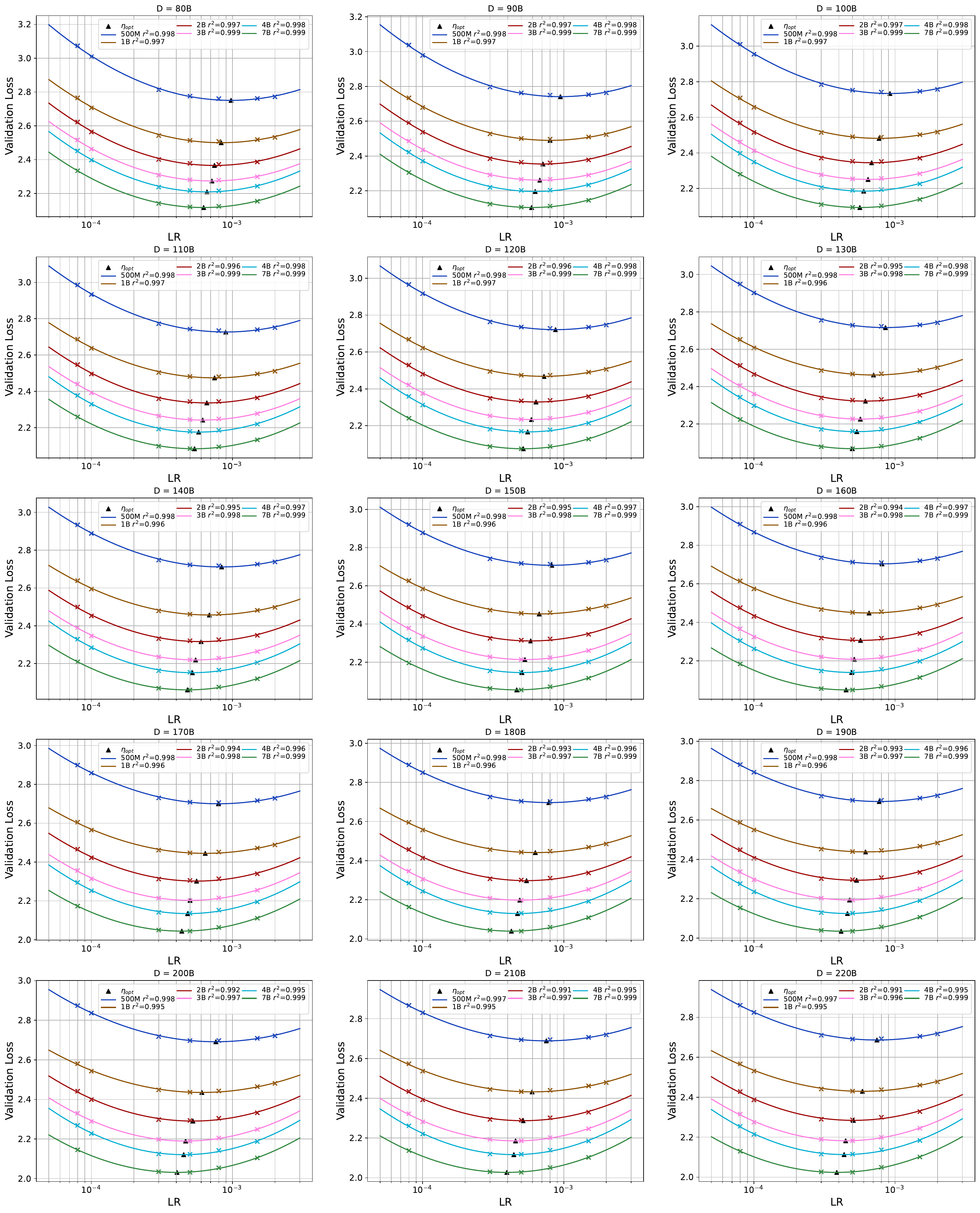}
    \caption{All results of fitting with Equation \ref{eq:loss-n-d} across different amount of data.}
    \label{fig:loss_lr_d}
\end{figure*}

We sample validation loss data points at 10B-token intervals, ranging from 80B to 220B tokens, to facilitate subsequent analysis and curve fitting. Figure \ref{fig:loss_lr_d_heatmap} illustrates the variation of validation loss as a function of the learning rate $\eta$ with different model size and data size. Figure \ref{fig:loss_lr_d_3d} presents a 3-D visualization of the relationship among loss, learning rate, and training data size. Upon observing a distinct local minimum, we employ the quadratic polynomial defined in Equation \ref{eq:loss-n-d} to fit the data (as shown in Figure \ref{fig:loss_lr_d}). The coefficients of determination ($R^2$) consistently exceed 0.995, enabling a precise estimation of the optimal learning rate based on these curves.

\begin{figure*}[h]
    \centering
    \vspace{-0.35cm}
    \subfigtopskip=2pt
    \subfigbottomskip=2pt
    \subfigcapskip=-5pt
    \subfigure[$\eta_{opt}$(D)]
    {
        \label{fig:minlr_d}
        \includegraphics[width=0.27\linewidth]{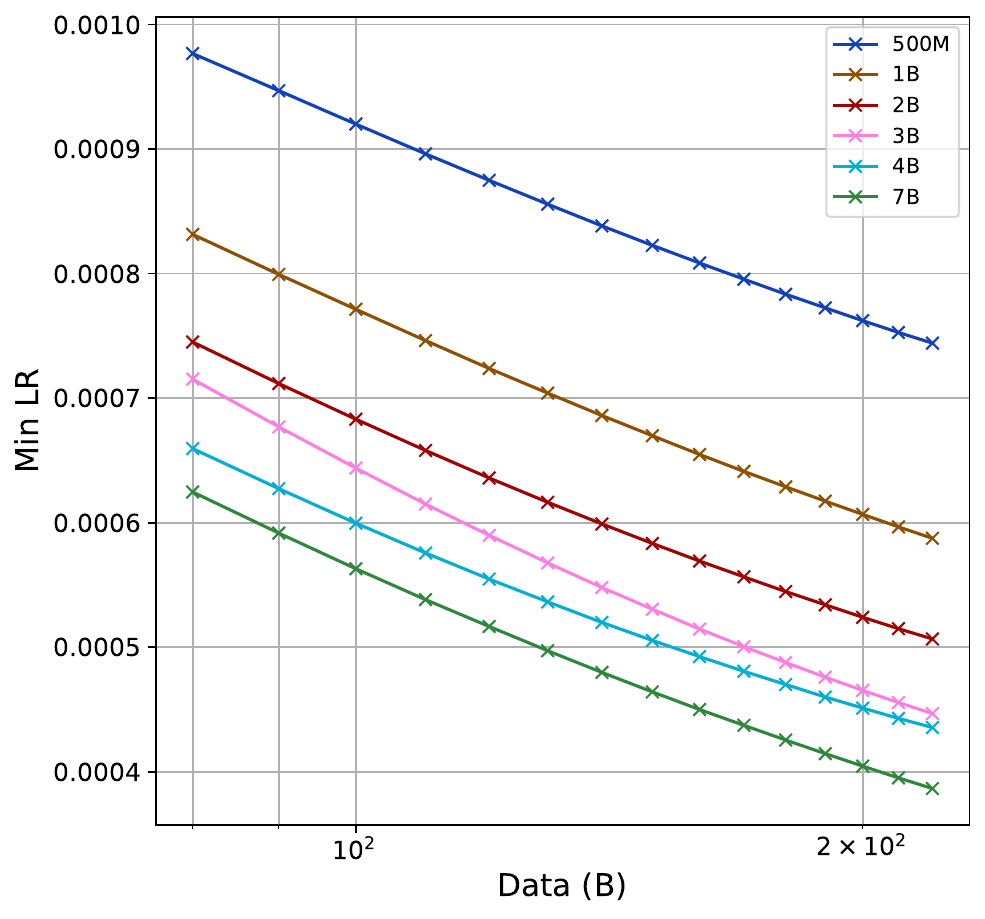}
    }
    \subfigure[$\eta_{opt}$(N)]
    {
        \label{fig:minlr_n}
        \includegraphics[width=0.27\linewidth]{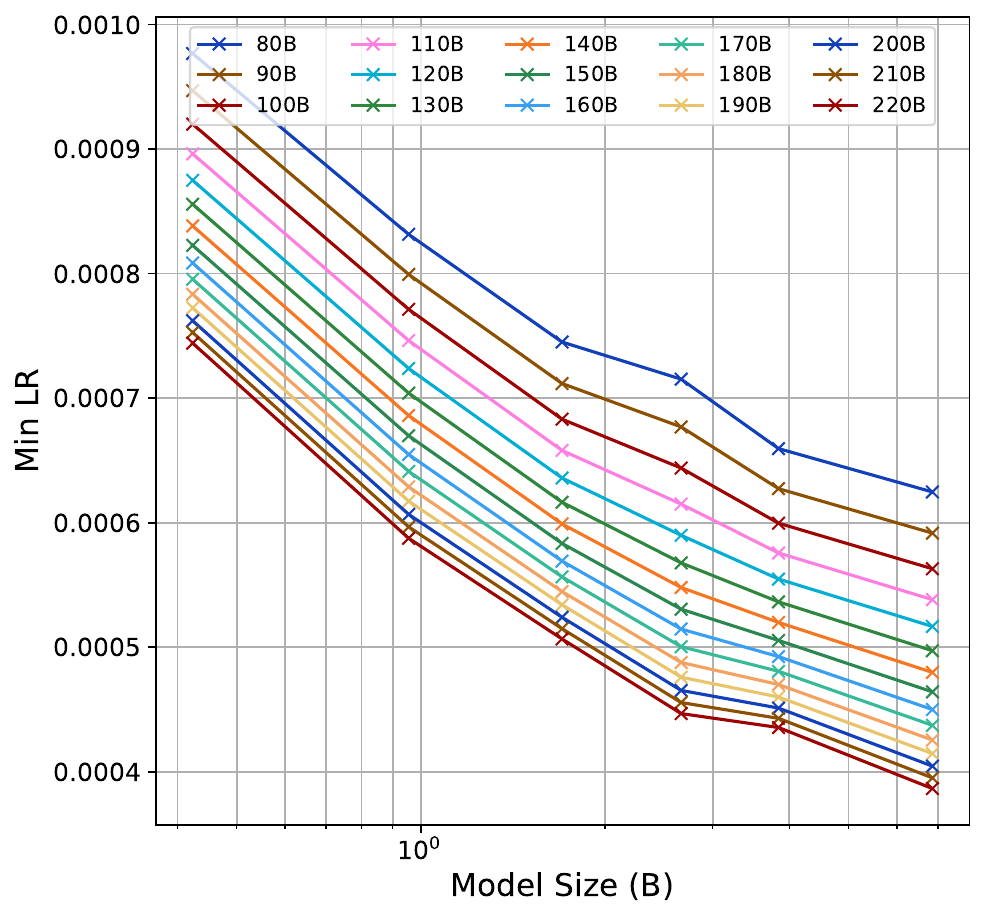}
    }
    \subfigure[$\eta_{opt}$(N, D)]
    {
        \label{fig:min_lr_heat}
        \includegraphics[width=0.37\linewidth]{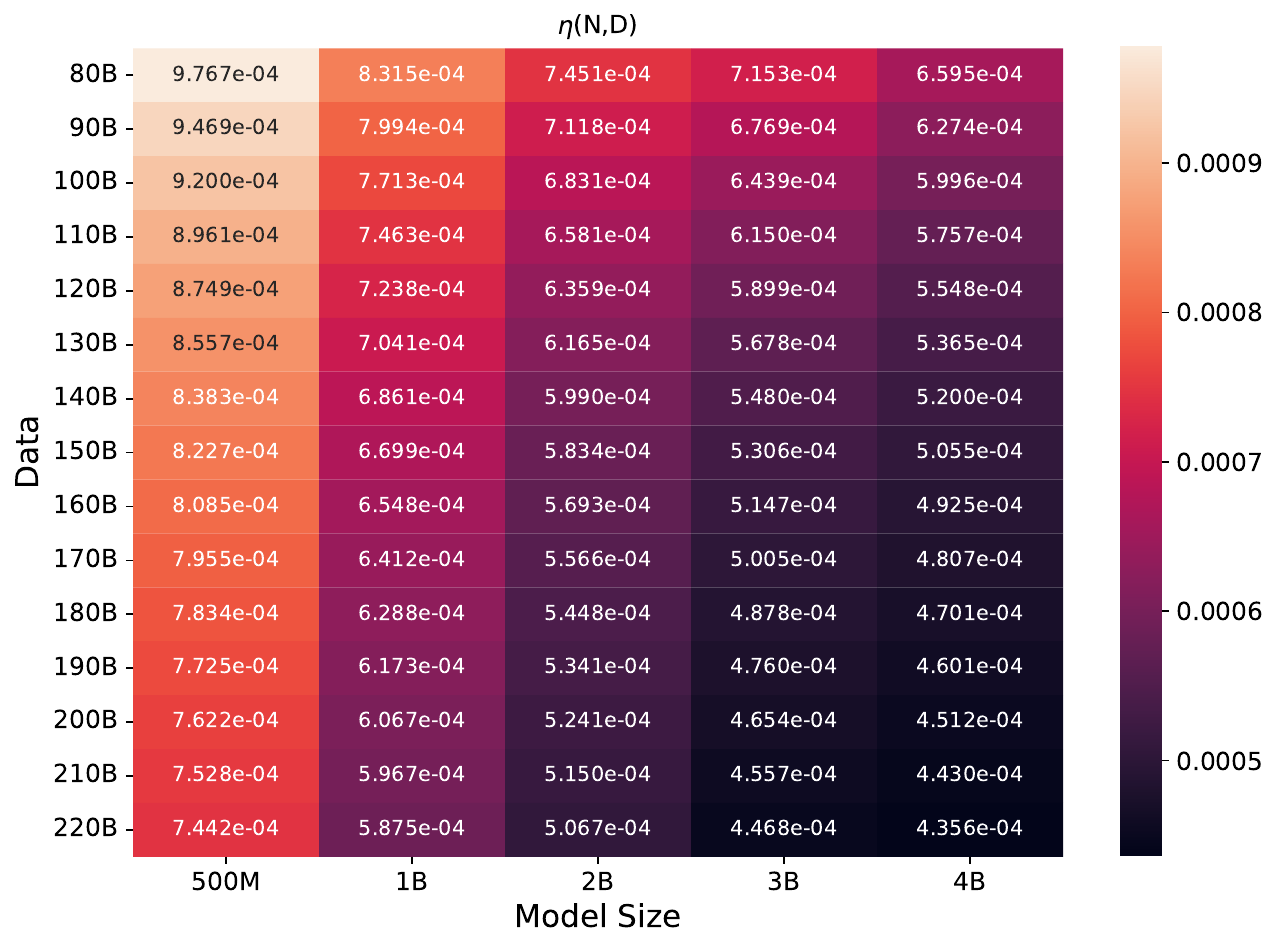}
    }
    \caption{Relationship ammog learning rate $\eta$, model size $N$ and training data size $D$.}
    \label{fig:min_lr}
\end{figure*}

As shown in Figure \ref{fig:minlr_d} and \ref{fig:minlr_n}, the global optimal learning rate exhibits a power-law relationship with both the model parameter count $N$ and training data size $D$. With reference to the studies of \citet{bjorck-scalinglaw}, we decide to use the following functional form:

\begin{equation} 
    \eta_{opt}(N, D) = C_{\eta} \cdot N^{-\alpha} \cdot D^{-\beta},
    \label{eq:lr-n-d-form}
\end{equation}

where $C_{\eta}, \alpha, \beta$ are positive constants. After employing non-linear least squares to fit the curve, we finally get the parameters of Equation \ref{eq:lr-n-d-res}:

\begin{equation}
    C_{\eta} \sim 38.4588, \alpha \sim 0.2219, \beta \sim 0.3509.
    \label{eq:lr-n-d-param}
\end{equation}

\subsubsection{Module-Level Optimal Learning Rates}
\label{sec:appendix-module-lr}

This subsection details the step-by-step process of searching Module-Level Optimal LR.

We split the model into the following four groups of parameters:

\begin{itemize}
    \item \textbf{Embedding Parameters}, which is the word embedding layer of a model,
    \item \textbf{Hidden Parameters}, mainly composed of self-attention and layer norm modules,
    \item \textbf{Router}, which contains the router matrix and experts,
    \item \textbf{LM Head Parameters}, which is the unembedding output layer.
\end{itemize}

Similar to our experiments in Section \ref{sec:appendix-module-lr}, while searching optimal LR across different module groups, the training data size in set to approximately 120B tokens. According to the results above, we can derive the global optimal LR $\eta_{opt}$ of every size of model in the experiment via Equation \ref{eq:loss-n-d}:

\begin{figure*}[t]
    \centering
    \includegraphics[width=1\linewidth]{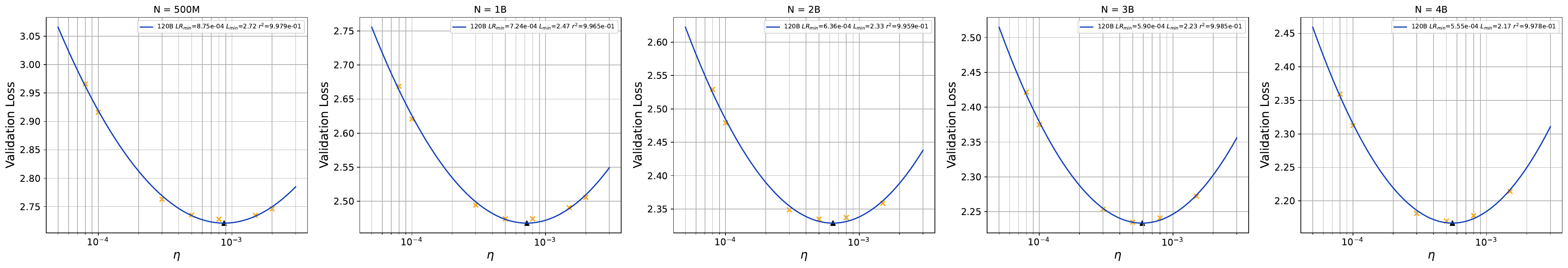}
    \caption{Initial stage of module-level learning rate searching.}
    \label{fig:module-lr-stage0}
\end{figure*}

\begin{table*}[htbp]
    \centering
    \caption{Global Optimal LR at 120B}
    \label{tab:module-lr-stage0}
    
    \begin{tabular}{lccccccccc}
        \toprule
        \textbf{$N$} & 
        \textbf{\shortstack{500M}} & 
        \textbf{\shortstack{1B}} & 
        \textbf{\shortstack{2B}} & 
        \textbf{\shortstack{3B}} & 
        \textbf{\shortstack{4B}} \\
        \midrule

        $\eta_{opt}$ & 8.75e-4 & 7.24e-4 & 6.36e-4 & 5.90e-4 & 5.55e-4 \\
        
        \bottomrule
    \end{tabular}
    \vspace{1ex}
\end{table*}

In the following stages, we sequentially conducting experiments in the order of LM Head, Router, Hidden, and Embedding parameters with greedy search strategy.

\begin{figure*}[t]
    \centering
    \includegraphics[width=1\linewidth]{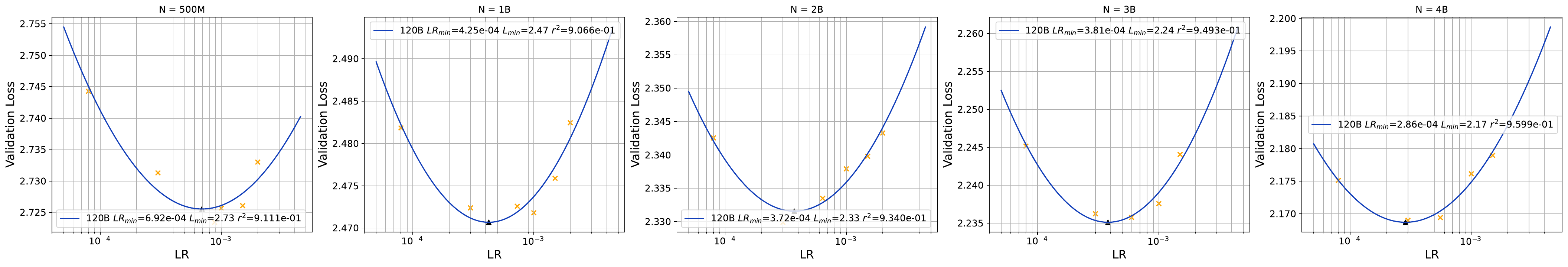}
    \caption{LM Head stage of module-level learning rate searching.}
    \label{fig:module-lr-stage1}
\end{figure*}

\textbf{LM Head}. First, we begin with the LM Head module. By varying the learning rate $\eta^{out}$ of the LM Head weights within a specified range while fixing the learning rates of all other weights to the current model’s global optimal learning rate(i.e. $\eta^{emb}=\eta^{hidden}=\eta^{router}=\eta_{opt}$), we conduct the search following the method described in Section \ref{sec:lr-scaling-law}. The curve fitted using Equation \ref{eq:loss-n-d} is shown in Figure \ref{fig:module-lr-stage1}, where the fitted minimum is taken as the module-level learning rate $\eta_{opt}^{out}$ for LM Head(Table \ref{tab:module-lr-stage1}).

\begin{table*}[htbp]
    \centering
    \caption{LM Head Optim LR at 120B}
    \label{tab:module-lr-stage1}
    
    \begin{tabular}{lccccccccc}
        \toprule
        \textbf{$N$} & 
        \textbf{\shortstack{500M}} & 
        \textbf{\shortstack{1B}} & 
        \textbf{\shortstack{2B}} & 
        \textbf{\shortstack{3B}} & 
        \textbf{\shortstack{4B}} \\
        \midrule

        $\eta_{opt}^{out}$ & 6.92e-4 & 425e-4 & 3.72e-4 & 3.81e-4 & 2.86e-4 \\
        
        \bottomrule
    \end{tabular}%
    \vspace{1ex}
\end{table*}

\begin{figure*}[t]
    \centering
    \includegraphics[width=1\linewidth]{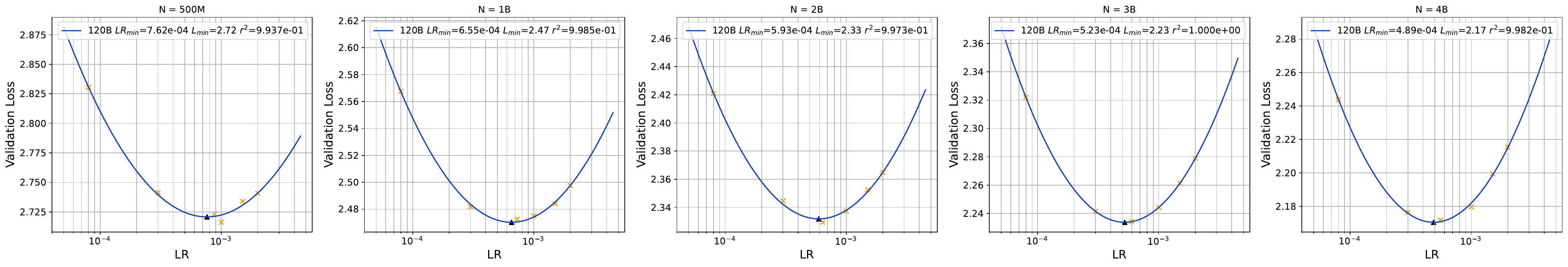}
    \caption{Router stage of module-level learning rate searching.}
    \label{fig:module-lr-stage2}
\end{figure*}

\textbf{Router}. In the second searching stage, we set $\eta^{emb}=\eta^{hidden}=\eta_{opt}, \eta^{out}=\eta_{opt}^{out}$ and search learning rate on Router layers. The results are illustrated in Figure \ref{fig:module-lr-stage2} and Table \ref{tab:module-lr-stage2}

\begin{table*}[htbp]
    \centering
    \caption{Router Optim LR at 120B}
    \label{tab:module-lr-stage2}
    
    \begin{tabular}{lccccccccc}
        \toprule
        \textbf{$N$} & 
        \textbf{\shortstack{500M}} & 
        \textbf{\shortstack{1B}} & 
        \textbf{\shortstack{2B}} & 
        \textbf{\shortstack{3B}} & 
        \textbf{\shortstack{4B}} \\
        \midrule

        $\eta_{opt}^{router}$ & 7.62e-4 & 6.55e-4 & 5.93e-4 & 5.23e-4 & 4.89e-4 \\
        
        \bottomrule
    \end{tabular}%
    \vspace{1ex}
\end{table*}

\begin{figure*}[t]
    \centering
    \includegraphics[width=1\linewidth]{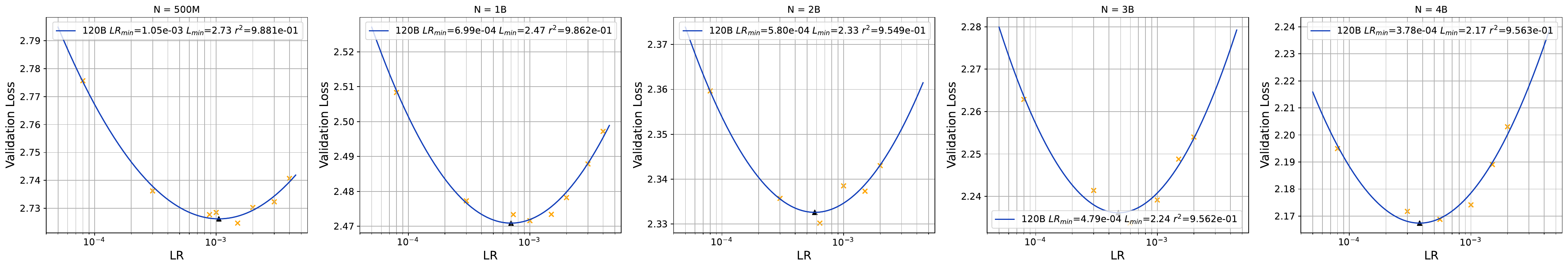}
    \caption{Hidden stage of module-level learning rate searching.}
    \label{fig:module-lr-stage3}
\end{figure*}

\textbf{Hidden}. Next, set $\eta^{emb}=\eta_{opt}, \eta^{router}=\eta_{opt}^{router}, \eta^{out}=\eta_{opt}^{out}$ while searching optimal learning rate on Hidden parameters to obtain $\eta_{opt}^{hidden}$. The results are illustrated in Figure \ref{fig:module-lr-stage3} and Table \ref{tab:module-lr-stage3}

\begin{table*}[htbp]
    \centering
    \caption{Hidden Optim LR at 120B}
    \label{tab:module-lr-stage3}
    
    \begin{tabular}{lccccccccc}
        \toprule
        \textbf{$N$} & 
        \textbf{\shortstack{500M}} & 
        \textbf{\shortstack{1B}} & 
        \textbf{\shortstack{2B}} & 
        \textbf{\shortstack{3B}} & 
        \textbf{\shortstack{4B}} \\
        \midrule

        $\eta_{opt}^{hidden}$ & 1.05e-3 & 6.99e-4 & 5.80e-4 & 4.79e-4 & 3.78e-4 \\
        
        \bottomrule
    \end{tabular}%
    \vspace{1ex}
\end{table*}

\begin{figure*}[t]
    \centering
    \includegraphics[width=1\linewidth]{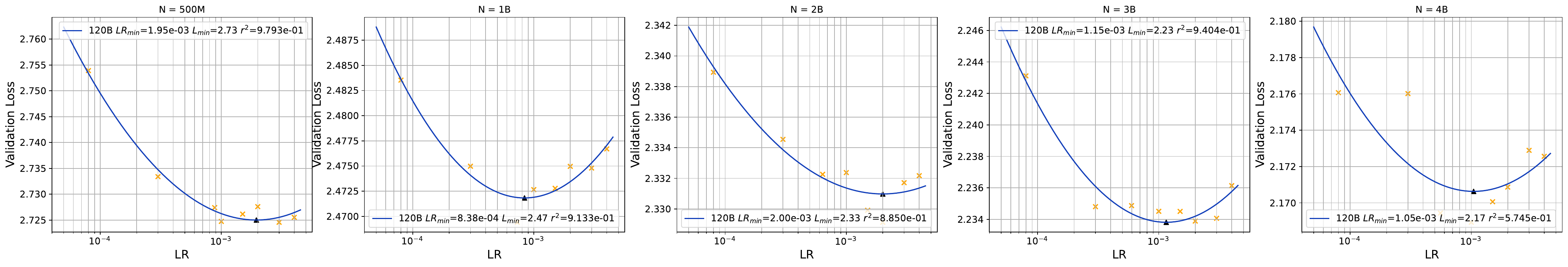}
    \caption{Embedding stage of module-level learning rate searching.}
    \label{fig:module-lr-stage4}
\end{figure*}

\textbf{Embedding}. Finally, we set $\eta^{router}=\eta_{opt}^{router}, \eta^{hidden}=\eta_{opt}^{hidden},\eta^{out}=\eta_{opt}^{out}$ and conduct learning rate searching on Embedding layer and get its optimal learning rate $\eta_{opt}^{emb}$. The results are illustrated in Figure \ref{fig:module-lr-stage4} and Table \ref{tab:module-lr-stage4}

\begin{table*}[htbp]
    \centering
    \caption{Embedding Optim LR at 120B}
    \label{tab:module-lr-stage4}
    
    \begin{tabular}{lccccccccc}
        \toprule
        \textbf{$N$} & 
        \textbf{\shortstack{500M}} & 
        \textbf{\shortstack{1B}} & 
        \textbf{\shortstack{2B}} & 
        \textbf{\shortstack{3B}} & 
        \textbf{\shortstack{4B}} \\
        \midrule

        $\eta_{opt}^{out}$ & 1.95e-3 & 8.38e-4 & 2.00e-3 & 1.15e-3 & 1.05e-3 \\
        
        \bottomrule
    \end{tabular}%
    \vspace{1ex}
\end{table*}

\textbf{Overall Results}. The overall results of module-level optimal learning rate are shown in Table \ref{tab:module-lr}

\begin{table*}[htbp]
    \centering
    \caption{Module-Level Optim LR at 120B}
    \label{tab:module-lr}
    
    \begin{tabular}{lccccccccc}
        \toprule
        \textbf{$N$} & 
        \textbf{\shortstack{500M}} & 
        \textbf{\shortstack{1B}} & 
        \textbf{\shortstack{2B}} & 
        \textbf{\shortstack{3B}} & 
        \textbf{\shortstack{4B}} \\
        \midrule

        $\eta_{opt}^{out}$ & 6.92e-4 & 425e-4 & 3.72e-4 & 3.81e-4 & 2.86e-4 \\
        $\eta_{opt}^{router}$ & 7.62e-4 & 6.55e-4 & 5.93e-4 & 5.23e-4 & 4.89e-4 \\
        $\eta_{opt}^{hidden}$ & 1.05e-3 & 6.99e-4 & 5.80e-4 & 4.79e-4 & 3.78e-4 \\
        $\eta_{opt}^{out}$ & 1.95e-3 & 8.38e-4 & 2.00e-3 & 1.15e-3 & 1.05e-3 \\
        
        \bottomrule
    \end{tabular}%
    \vspace{1ex}
\end{table*}

\subsubsection{$\mu$Transfer}

We refer to \citet{complete-d-p} to conduct our $\mu$Transfer experiments. The transfer method is shown in Table \ref{tab:mutransfer}. As we maintain an invariant batch size across all experimental configurations, we only consider the influence of training token counts alongside model width and depth when employ $\mu$Transfer.

\begin{table*}[htbp]
\centering
\small
\caption{Hyperparameters' transfer rule of $\mu$Transfer}
\label{tab:mutransfer}
\vspace{2mm}
\begin{tabular}{@{}llcccc@{}}
\toprule
& \textbf{Parameterisation:} & & $\mu$\textbf{P} & \multicolumn{2}{c}{\textbf{Complete}$^{(d)}$\textbf{P}} \\ \midrule
\multirow{3}{*}{Multipliers} & MHA Residual & & $\mathbf{x} + \text{MHABlock}(\mathbf{x})$ & $\mathbf{x} + m_L^{-\alpha} \text{MHABlock}(\mathbf{x})$ \\
 & MLP Residual & & $\mathbf{x} + \text{MLPBlock}(\mathbf{x})$ & $\mathbf{x} + m_L^{-\alpha} \text{MLPBlock}(\mathbf{x})$ \\ 
 & Unemb. Fwd & & Unaugmented & Unaugmented &  \\ \midrule
\multirow{5}{*}{\shortstack{Init Variances}} & Input Emb. & \multirow{5}{*}{$\sigma_b^2$ } & & & \\
 & Hidden weights & & $\times m_N^{-1}$ & $\times m_N^{-1}$ & \\
 & Hidden biases/norms & & & & \\
 & Unemb. LN & & & & \\
 & Unemb. Weights & & $\times m_N^{-2}$ & $\times m_N^{-2}$ & \\ \midrule
\multirow{5}{*}{\shortstack{Learning Rates}} & Input Emb. & \multirow{5}{*}{$\eta_b$} & & & \multirow{5}{*}{$\times \sqrt{\frac{1}{m_D}}$}  \\
 & Hidden weights & & $\times m_N^{-1}$ & $\times m_N^{-1} \times {\color{blue}m_L^{\alpha-1}}$ & \\
 & Hidden biases/norm & & & ${\color{blue}\times m_L^{\alpha-1}}$ & \\
 & Unemb. LN & & & & \\
 & Unemb. weights & & $\times m_N^{-1}$ & $\times m_N^{-1} \quad$ & \\ \midrule
\multirow{4}{*}{\shortstack{AdamW $\epsilon$}} & Hidden weights/biases/norms & \multirow{4}{*}{$\epsilon_b$} & $\times m_N^{-1}$ & $\times m_N^{-1} \times {\color{blue}m_L^{-\alpha}}$ & \multirow{4}{*}{$\times \sqrt{m_D}$} \\
 & QK norms & & NA & ${\color{blue}\times m_L^{-\alpha}}$  & \\
 & Input Emb. & & $\times m_N^{-1}$ & $\times m_N^{-1} $ & \\
 & Output weights/biases/norms & & & \\ \midrule
\multirow{3}{*}{\shortstack{Weight decay}} & Hidden weights &\multirow{3}{*}{$\lambda_b$} & $\times m_N$ & $\times m_N$ & \multirow{3}{*}{$\times \sqrt{\frac{1}{m_D}}$} \\
 & Unemb. weights & & $\times m_N$ & $\times m_N$ & \\
 & Rest & & $\times 1$ & $\times 1$ & \\ \midrule
\end{tabular}
\end{table*}

\end{document}